\PassOptionsToPackage{table,xcdraw}{xcolor}

\documentclass[sigconf]{acmart}

\newcommand{\methodname}{{\tt{FedLoRA}}}
\usepackage{subcaption}
\usepackage[ruled]{algorithm2e}
\SetKwInput{KwInput}{Input}                
\SetKwInput{KwOutput}{Output}              
\definecolor{ylp_color1}{RGB}{255,193,193}
\definecolor{ylp_color2}{RGB}{255,228,225}
\usepackage{tcolorbox}
\newtcbox{\mybox}[1][red]{on line, colback = {RGB}{255,228,225}, colframe = {RGB}{255,193,193},  arc=1mm, auto outer arc, boxrule=0.5pt,}
\usepackage{multirow}
\usepackage{booktabs}
\usepackage{wrapfig}
\usepackage{amsmath}
\usepackage{amsthm}
\usepackage{amsfonts}
\usepackage{bm}
\usepackage{eucal}
\usepackage{arydshln}

\newtheorem{theorem}{Theorem}[section]

\newtheorem{lemma}[theorem]{Lemma}

\newtheorem{assumption}[theorem]{Assumption}

\usepackage{pifont}
\usepackage{setspace}
\usepackage{url}

\AtBeginDocument{%
  \providecommand\BibTeX{{%
    \normalfont B\kern-0.5em{\scshape i\kern-0.25em b}\kern-0.8em\TeX}}}

\setcopyright{acmcopyright}
\copyrightyear{2023}
\acmYear{2023}
\acmDOI{XXXXXXX.XXXXXXX}

\acmConference[Conference XX]{Make sure to enter the correct
  conference title from your rights confirmation emai}{XX}{XX}
\acmPrice{15.00}
\acmISBN{XX}

\setcopyright{none}
\renewcommand\footnotetextcopyrightpermission[1]{}
\begin{document}

\title{pFedLoRA: Model-Heterogeneous Personalized Federated Learning with LoRA Tuning}

\author{Liping Yi}
\affiliation{%
  \institution{College of C.S., TMCC, SysNet, DISSec, GTIISC \\ Nankai University}
  \city{Tianjin}
  \country{China}
}
\email{yiliping@nbjl.nankai.edu.cn}

\author{Han~Yu}
\authornote{Corresponding authors.}
\affiliation{%
  \institution{School of Computer Science and Engineering \\ Nanyang Technological University}
  \country{Singapore}
}
\email{han.yu@ntu.edu.sg}

\author{Gang Wang}
\authornotemark[1]
\affiliation{%
  \institution{College of C.S., TMCC, SysNet, DISSec, GTIISC \\ Nankai University}
  \city{Tianjin}
  \country{China}
}
\email{wgzwp@nbjl.nankai.edu.cn}

\author{Xiaoguang Liu}
\affiliation{%
  \institution{College of C.S., TMCC, SysNet, DISSec, GTIISC \\ Nankai University}
  \city{Tianjin}
  \country{China}
}
\email{liuxg@nbjl.nankai.edu.cn}

\author{Xiaoxiao Li}
\email{xiaoxiao.li@ece.ubc.ca}
\affiliation{%
  \institution{Electrical and Computer Engineering Department, University of British Columbia (UBC)}
  \city{Vancouver}
  \country{Canada}
}

\renewcommand{\shortauthors}{L. Yi and H. Yu, et al.}

\begin{abstract}
Federated learning (FL) is an emerging machine learning paradigm in which a central server coordinates multiple participants (clients) collaboratively to train on decentralized data. In practice, FL often faces statistical, system, and model heterogeneities, which inspires the field of Model-Heterogeneous Personalized Federated Learning (MHPFL). With the increased interest in adopting large language models (LLMs) in FL, the existing MHPFL methods cannot achieve acceptable computational and communication costs, while maintaining satisfactory model performance. 
To bridge this gap, we propose a novel and efficient model-heterogeneous \underline{p}ersonalized \underline{Fed}erated learning framework based on \underline{LoRA} tuning (\methodname{}). Inspired by the popular LoRA method for fine-tuning pre-trained LLMs with a low-rank model (\emph{a.k.a}., an adapter), we design \textit{a homogeneous small adapter} to facilitate federated \textit{client's heterogeneous local model} training with our proposed \textit{iterative training} for global-local knowledge exchange. The homogeneous small local adapters are aggregated on the FL server to generate a global adapter.
We theoretically prove the convergence of \methodname{}. Extensive experiments on two benchmark datasets demonstrate that \methodname{} outperforms six state-of-the-art baselines, beating the best method by $1.35\%$ in test accuracy, $11.81 \times$ computation overhead reduction and $7.41\times$ communication cost saving.
\end{abstract}

\maketitle

\section{Introduction}
As data privacy laws such as GDPR \cite{1w-survey} have been rolled out worldwide due to concerns about privacy leakage, the traditional machine learning paradigm relying on collecting data for model training faces increasing challenges.
Federated learning (FL) \cite{FedAvg} has emerged as a collaborative learning paradigm in response to such a trend. In a typical FL system, a central FL server broadcasts a global model to clients, who then train it on local data and upload the resulting model back to the server. The server aggregates the received local models to update the global model. These steps repeat until the global model converges. Only models are transmitted between the server and clients without exposing private local data.

The above design requires that all clients have to train models with the same structures (homogeneous), which makes the traditional FL paradigm unsuitable when facing various types of heterogeneity \cite{yi2023fedgh}:
    \textbf{Statistical (Data) Heterogeneity}. FL clients' local data often follow non-independent and identical distributions (non-IID). A local model solely trained by the client might perform better than the global FL model trained on non-IID data. 
    \textbf{Resource Heterogeneity}. Clients participating in FL can be mobile edge devices \cite{PruneFL} with different hardware resources (\emph{e.g.} computation power and bandwidth). 
    Traditional FL requires all (resource heterogeneous) clients to train models with the same structures, leading to model performance bottleneck as low-resource clients can only support smaller models.
    \textbf{Model Heterogeneity}. When FL participants are enterprises, they often maintain private model repositories with heterogeneous models. Fine-tuning them during FL training not only saves training time but also protects intellectual property \cite{MHPFL-survey}.  

These challenges motivate the research field of Model Heterogeneous Personalized Federated Learning (MHPFL). Existing MHPFL methods can be divided into three categories: 1) \textit{Knowledge distillation-based MHPFL} methods \cite{FedMD,FedDF} often rely on a public dataset with the same distribution as local data, but such suitable public datasets may not always be available. Other knowledge distillation-based MHPFL methods \cite{FD,FedProto} without requiring public datasets often incur high computation and/or communication costs for FL clients due to local distillation. 2) \textit{Model mixup-based MHPFL} methods \cite{LG-FedAvg,FedRep} split each local model into a heterogeneous part for local training and a homogeneous part for model aggregation. Only parts of the entire model are aggregated by the server results heterogeneous local models often suffer from subpar performance. 3) \textit{Mutual learning-based MHPFL} methods \cite{FML,FedKD} assign a large heterogeneous model and a small homogeneous model for each client. The two models are trained locally via mutual learning, and only the small homogeneous model is uploaded to the server for aggregation. Training two models locally incurs extra computation costs. Moreover, the undefined choice of the model structures also affects the resulting performance.

\begin{figure}[!t]
\centerline{\includegraphics[width=0.65\linewidth]{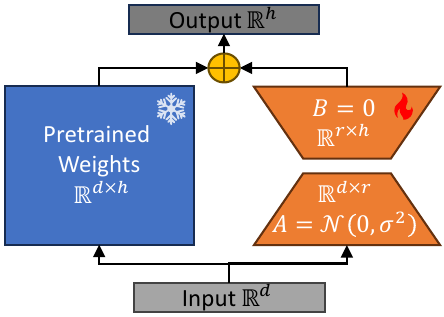}}
\vspace{-0.5em}
\caption{The working principle of LoRA.}\label{fig:LoRA}
\vspace{-1em}
\end{figure}

Low-Rank Adaptation (LoRA) \cite{LoRA} has recently emerged as a popular method for fine-tuning pre-trained large language models (LLMs) to fit downstream tasks. As shown in Figure~\ref{fig:LoRA}, it adds a branch alongside the pre-trained model, which is a low-rank adapter with the same input sample and the same output dimension as the pre-trained model. During fine-tuning, it freezes the pre-trained large model and trains the low-rank adapter. In each iteration of training, one sample is input into the frozen pre-trained model and the training adapter simultaneously, and the outputs of this sample from two models are summed as the final output. Then the hard loss between the final output and the label is calculated for updating the adapter by gradient descent. After that, the combination of the two branches is used for model inference, which may perform a similar or higher accuracy than fine-tuning the whole pre-trained model directly. As only the small adapter is trained during fine-tuning, LoRA achieves efficient computation and storage.

Inspired by LoRA, we propose an efficient model-heterogeneous \underline{p}ersonalized \underline{Fed}rated learning framework based on \underline{LoRA} tuning (\methodname{}) for supervised learning tasks. Belonging to the mutual learning category, each client holds heterogeneous data and model. \methodname{} enables \textit{a small low-rank homogeneous adapter} to be incorporated into the large heterogeneous local model.
In each communication round, 1) clients first replace their local adapters with the global homogeneous adapter received from the FL server; 2) then, they perform the proposed \textit{iterative learning} method to train the two models alternatively for global-local knowledge transfer; and 3) finally, the updated local homogeneous adapters from the clients are aggregated by the FL server.
In short, adapters are regarded as ``knowledge carriers'' for aggregation to support knowledge transfer among clients.
Each client only additionally trains an small adapter and only communicate small homogeneous adapters with the server. Such a design ensures that \methodname{} achieves MHPFL efficiently.

As the insertion of LoRA adapters changes the process of local model training compared with traditional FL, we derive the non-convex convergence rate of \methodname{} based on local iterative training and prove that it converges over time. Extensive experiments on two benchmark datasets demonstrate significant advantages of \methodname{} in both model-homogeneous and model-heterogeneous scenarios compared to six state-of-the-art methods, beating the best of them by $1.35\%$ in test accuracy, $11.81 \times$ computation overhead reduction and $7.41\times$ communication cost saving.





\section{Related Work}\label{sec:related}
Existing MHPFL methods have two branches: a) \textit{Partially model-heterogeneous}, clients hold different subnets of the global model, and heterogeneous subnets can be aggregated on the server, such as {\tt{FedRolex}} \cite{FedRolex}, {\tt{HeteroFL}} \cite{HeteroFL}, {\tt{FjORD}} \cite{FjORD}, {\tt{HFL}} \cite{HFL}, {\tt{Fed2}} \cite{Fed2}, {\tt{FedResCuE}} \cite{FedResCuE}. b) \textit{Completely model-heterogeneous}, clients hold models with completely different model structures that can not be aggregated directly on the server. This branch can be further divided into the following categories. 


\textbf{Knowledge Distillation-based MHPFL}. In the \textit{public dataset-dependent} knowledge distillation-based MHPFL methods (such as {\tt{Cronus}} \cite{Cronus}, {\tt{FedGEMS}} \cite{FedGEMS}, {\tt{Fed-ET}} \cite{Fed-ET}, {\tt{FSFL}} \cite{FSFL}, {\tt{FCCL}} \cite{FCCL}, {\tt{DS-FL}} \cite{DS-FL}, {\tt{FedMD}} \cite{FedMD}, {\tt{FedKT}} \cite{FedKT}, {\tt{FedDF}} \cite{FedDF},   {\tt{FedHeNN}} \cite{FedHeNN},   {\tt{FedAUX}} \cite{FEDAUX},  {\tt{CFD}} \cite{CFD}, {\tt{FedKEMF}} \cite{FedKEMF},   {\tt{KT-pFL}} \cite{KT-pFL}), 
the server aggregates the output logits of different clients' heterogeneous models on a public dataset to construct global logits. But the public dataset is not always accessible and the algorithm performs well only if the public dataset has the same distribution as private data. Besides, transmitting logits of \textit{each} public data sample incurs high communication costs for a large-scale public dataset.
For other knowledge distillation-based MHPFL methods not dependent on a public dataset, {\tt{FedZKT}} \cite{FedZKT} and {\tt{FedGen}} \cite{FedGen} introduce zero-shot knowledge distillation to FL, they generate a public dataset through training a generator, which is time-consuming. {\tt{HFD}} \cite{HFD1,HFD2}, {\tt{FedGKT}} \cite{FedGKT}, {\tt{FD}} \cite{FD},  {\tt{FedProto}} \cite{FedProto}  
allow each client to upload the local (average) logits or representations of its seen-class samples to the server for aggregation by class to generate the global class-logits or representations which are sent back to clients and used to calculate the distillation loss with local logits for each local data sample, incurring high computational overheads.

\textbf{Model Mixup-based MHPFL}. These methods split each client's local model into two parts: one feature extractor and one classifier, and only one part is shared. {\tt{FedMatch}} \cite{FedMatch}, {\tt{FedRep}} \cite{FedRep}, {\tt{FedBABU}} \cite{FedBABU} and {\tt{FedAlt/FedSim}} \cite{FedAlt/FedSim} share homogeneous feature extractors to enhance model generalization and personalize local classifier. In contrast, {\tt{FedClassAvg}} \cite{FedClassAvg},  {\tt{LG-FedAvg}} \cite{LG-FedAvg} and {\tt{CHFL}} \cite{CHFL} share homogeneous classifier to improve model classification and personalize local feature extractor. Since only partial parameters of the whole model are shared, the final local heterogeneous models face performance bottlenecks.

\textbf{Mutual Learning-based MHPFL}. {\tt{FML}} \cite{FML} and {\tt{FedKD}} \cite{FedKD} assign a small homogeneous model and a large heterogeneous model in each client, and train them in a \textit{mutual learning} manner. The small homogeneous models after local training are aggregated on the server. In short, the small homogeneous models as information mediums implement the knowledge transfer across large heterogeneous models. However, they do not explore the relationship between the model structure and parameter capacity between the two models, which may affect the final model performance and computation costs of training an extra small homogeneous model for each client.


\textbf{Our Insight}. \methodname{} enables knowledge transfer across clients' local large heterogeneous models through \textit{a small low-rank homogeneous adapter}, which is a low-rank version of the fully connected layers of the large local heterogeneous models. It does not rely on any public dataset, and incurs low computation and communication costs as only small low-rank extra adapters are trained on clients and transmitted between the FL server and the clients.

\section{Preliminaries}


\subsection{LoRA Adapter}\label{sec:lora-adapter}
As shown in Figure.~\ref{fig:LoRA}, a LoRA adapter has the same input and output dimensions as the pre-trained large model. As elaborated in \cite{LoRA}, the structures of the current LoRA support linear, embedding and convolutional layers. Take a linear LoRA as an example. Given a linear layer of the pre-trained model with $\mathbb{R}^{d}$ input and $\mathbb{R}^{h}$ output (\emph{i.e.}, with a $\mathbb{R}^{d \times h}$ parameter matrix), a linear LoRA adapter can be a combination of two small matrices $A (\mathbb{R}^{d \times r})$ and $B (\mathbb{R}^{r \times h})$ by \textit{matrix decomposition}, where the rank $r$ is far smaller than $d$ and $h$.
Before training, matrix $A$ can be initialized with a Gaussian distribution $\mathcal{N}{(0, \sigma^2)}$ ($0$: mean, $\sigma^2$: variance), and matrix $B$ can be initialized with $0$.

\subsection{Overview of Federated Learning}
{\tt{FedAvg}} \cite{FedAvg} is a typical FL algorithm, it assumes that a FL system consists of one central server and $N$ clients. In each communication round, the server randomly selects a fraction $C$ of clients $S$ ($|S|= \lfloor C N \rfloor = K$) and broadcasts the global model $\mathcal{F}(\omega)$ ($\mathcal{F}(\cdot)$ is model structure, $\omega$ are model parameters) to the selected $K$ clients. Client $k$ trains the received global model $\mathcal{F}(\omega)$ on its local data $D_k$ ($D_k \sim P_k$, local data $D_k$ obeys distribution $P_k$, \emph{i.e.}, local data from different clients are non-IID) to obtain updated local model $\mathcal{F}(\omega_k)$ by gradient descent, \emph{i.e.}, $\omega_k \gets \omega-\eta \nabla \ell(\mathcal{F}(\boldsymbol{x}_i; \omega), y_i)$. $\ell(\mathcal{F}(\boldsymbol{x}_i; \omega), y_i)$ is the loss of the global model $\mathcal{F}(\omega)$ on the sample $(\boldsymbol{x}_i,y_i) \in D_k$. The updated local model $\mathcal{F}(\omega_k)$ is uploaded to the server. The server aggregates the received local models from the selected $K$ clients by weighted averaging to update the global model, \emph{i.e.}, $\omega=\sum_{k=0}^{K-1} \frac{n_k}{n} \omega_k$ ($n_k=|D_k|$ is data volume of client $k$, $n=\sum_{k=0}^{N-1} n_k$ is data volume of all clients).

In short, the typical FL algorithm requires all clients to train local models with the same structures (\textbf{homogeneous}), and its training objective is to minimize the average loss of the global model $\mathcal{F}(\omega)$ on all client data, \emph{i.e.},
\begin{equation}
\min _{\omega \in \mathbb{R}^d} \sum_{k=0}^{K-1} \frac{n_k}{n} \mathcal{L}_k(D_k ; \mathcal{F}(\omega)),
\end{equation}
where the parameters $\omega$ of the global model are $d$-dimensional real numbers, $\mathcal{L}_k(D_k ; \mathcal{F}(\omega))$ is the average loss of the global model $\mathcal{F}(\omega)$ on client $k$'s local data $D_k$.

\subsection{Problem Definition}
The goal of this paper is to study model-heterogeneous personalized FL in supervised image classification tasks. We assume that all clients execute the same image classification task, and different clients may hold \textbf{heterogeneous} local models $\mathcal{F}_{k}(\omega_k)$ ($\mathcal{F}_{k}(\cdot)$ denotes different model structures, $\omega_k$ indicates personalized model parameters).

\begin{figure}[b]
\centerline{\includegraphics[width=0.9\linewidth]{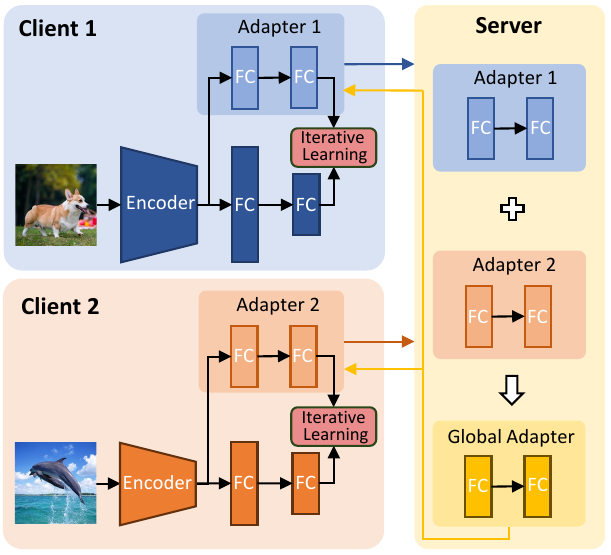}}
\vspace{-0.5em}
\caption{Workflow of \methodname{}.}
\label{fig:FedLoRA-workflow}
\vspace{-1em}
\end{figure}

To support generalized knowledge exchanging in FL training involving heterogeneous local models, we insert a small low-rank homogeneous adapter $\mathcal{A}(\theta_k)$ ($\mathcal{A}(\cdot)$ is adapter structure, $\theta_k$ are personalized local adapter parameters) into a large local heterogeneous model $\mathcal{F}_{k}(\omega_k)$. Clients share the small low-rank homogeneous adapters to implement the knowledge transfer across heterogeneous models from different clients. As shown in Figure~\ref{fig:FedLoRA-workflow}, the model consisting of the small low-rank homogeneous adapter and the large heterogeneous model is denoted as $\mathcal{F}_{k}(\omega_k) + \mathcal{A}(\theta_k)$. The objective of \methodname{} is to minimize the sum of the loss of all clients' heterogeneous models, \emph{i.e.},
\begin{equation}
\min _{\omega_0, \ldots, \omega_{K-1} \in \mathbb{R}^{d_0, \ldots, d_{K-1}}} \sum_{k=0}^{K-1} \mathcal{L}_k(D_k ; \mathcal{F}_k(\omega_k)+\mathcal{A}(\theta_k)),
\end{equation}
where the parameters $\omega_0, \ldots, \omega_{K-1}$ of local heterogeneous models are $d_0, \ldots, d_{K-1}$-dimensional real numbers.

\section{The Proposed \methodname{} Approach}
To reduce the computational overhead incurred by training clients' low-rank adapters, unlike LoRA which matches an adapter for the entire pre-trained model, we view each client's personalized heterogeneous model as two parts: 1) the convolutional layers $f_{k}(\omega_{k,conv})$, and 2) the fully-connected layers $h_{k}(\omega_{k,fc})$, i.e, $\mathcal{F}_{k}(\omega_{k})=f_{k}(\omega_{k,conv}) \circ h_{k}(\omega_{k,fc})$, and we only insert a low-rank adapter $\mathcal{A}(\theta_k)$ for each client's fully-connected layers $h_{k}(\omega_{k,fc})$, as shown in Figure~\ref{fig:FedLoRA-workflow}.
The workflow of \methodname{} is as follows:

\begin{itemize}
\setlength{\itemsep}{0pt}
\setlength{\parsep}{0pt}
\setlength{\parskip}{0pt}
    \item In the $t$-th communication round, the server broadcasts the global low-rank adapter $\mathcal{A}(\theta^{t-1})$ to randomly selected $K$ clients. Client $k$ replaces its local adapter $\mathcal{A}(\theta^{t-1}_k)$ with the received global adapter $\mathcal{A}(\theta^{t-1})$.

    \item During local training, if we train the heterogeneous local model and the homogeneous adapter synchronously like LoRA, \emph{i.e.}, summing the output of two models for loss calculation, the immature global adapter in the beginning communication rounds of FL may lead to poor model performances. To boost model accuracy, we devise a novel \textbf{iterative learning} manner to train the two models for global-local knowledge transfer.

    \item After local iterative learning, the updated heterogeneous local models are stored in clients and the updated homogeneous local adapters are uploaded to the server for aggregation like {\tt{FedAvg}} to update the global adapter $\mathcal{A}(\theta^t)$, which fuses the knowledge from different clients' heterogeneous local models.
\end{itemize}


The above steps repeat until all personalized heterogeneous local models converge, which will be used for \textbf{inference} after federated training. More detailed description of \methodname{} is given in Algorithm~\ref{alg:FedLoRA} (Appendix~\ref{app:alg}).

\subsection{Iterative Learning}

\begin{figure}[t]
\centerline{\includegraphics[width=0.8\linewidth]{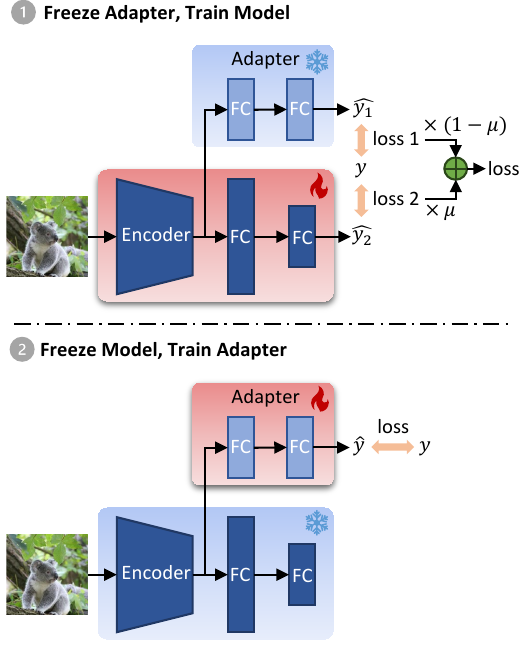}}
\caption{Iterative learning in \methodname{}.}
\label{fig:FedLoRA-iterative}
\vspace{-1em}
\end{figure}

Treating the local heterogeneous model and the homogeneous low-rank adapter as parts of a whole local model and training them \textit{simultaneously} is intuitive. However, training such a larger model might slow convergence and even lead to model performance degradation if local data are limited.
To boost the performance of personalized heterogeneous local models, we propose an \textit{iterative learning} method to train the heterogeneous local models and the homogeneous low-rank adapters. As illustrated in Figure~\ref{fig:FedLoRA-iterative}, firstly, we freeze the global adapter received by clients and train heterogeneous local models, which transfer global knowledge to clients. Then, we freeze the updated heterogeneous local models and train homogeneous low-rank adapters which are uploaded to the server for aggregation, which transfers local knowledge to the FL server.

\textbf{Freeze Adapter, Train Local Model}.
As Step \ding{192} shown in Figure~\ref{fig:FedLoRA-iterative}, client $k$ inputs the sample $(\boldsymbol{x}, y) \in D_k$ into the encoder (convolutional layers $f_{k}(\omega_{k,conv}^{t-1})$) of the local heterogeneous model to obtain representation $\boldsymbol{\mathcal{R}}=f_{k}\left(\boldsymbol{x};\omega_{k,conv}^{t-1}\right)$. Then, the representation $\boldsymbol{\mathcal{R}}$ is fed into the fully-connected layers $h_{k}(\omega_{k,fc}^{t-1})$ of the heterogeneous local model and the low-rank adapter $\mathcal{A}(\theta^{t-1})$ to obtain
\begin{equation}
    \widehat{y_1}=\mathcal{A}({\boldsymbol{\mathcal{R}};\theta}^{t-1}),
    \widehat{y_2}=h_k({\boldsymbol{\mathcal{R}};\omega}_{k,fc}^{t-1}).
\end{equation}
Then, the hard loss (such as cross-entropy loss \cite{CEloss}) between the output prediction $\widehat{y_1}$ of the homogeneous adapter and label $y$, and the hard loss between the output prediction $\widehat{y_2}$ of the heterogeneous local model and label $y$ can be calculated, respectively, \emph{i.e.},
\begin{equation}
    \ell_1=\ell(\widehat{y_1},y),\ \ell_2=\ell(\widehat{y_2},y).
\end{equation}

In the beginning communication rounds, the immature global adapter may have a negative influence on the performances of heterogeneous local models. To balance the global knowledge carried by the global adapter and the personalized local knowledge incorporated in the fully connected layers of local heterogeneous models, we take the linearly weighted sum of the hard losses from the two branches as the complete loss on the input sample, \emph{i.e.},
\begin{equation}\label{eq:miu}
    \ell_\omega=(1-\mu)\cdot\ell_1+\mu\cdot\ell_2,\ \mu\in[0.5,1).
\end{equation}
Then, we use the complete loss to update the heterogeneous local models by gradient descent (\emph{e.g.} SGD \cite{SGD}),
\begin{equation}
    \omega_k^t\gets\omega_k^{t-1}-\eta_\omega\nabla\ell_\omega,
\end{equation}
where $\eta_\omega$ is the learning rate of the heterogeneous local model. During this training process, the global knowledge carried by the frozen global adapter is transferred to heterogeneous local models, which promotes the generalization improvements of heterogeneous local models. Meanwhile, the personalized local knowledge involved in local data is learned by heterogeneous local models further, which facilitates the personalization of heterogeneous local models.


\textbf{Freeze Local Model, Train Adapter}. As Step \ding{193} shown in Figure~\ref{fig:FedLoRA-iterative}, client $k$ inputs the sample $(\boldsymbol{x}, y) \in D_k$ into the encoder (convolutional layers $f_{k}(\omega_{k,conv}^{t})$) of the \textit{ updated} local heterogeneous model to obtain representation $\widetilde{\boldsymbol{\mathcal{R}}}=f_{k}(\boldsymbol{x};\omega_{k,conv}^{t})$, then the representation $\widetilde{\boldsymbol{\mathcal{R}}}$ is input into the 
adapter $\mathcal{A}(\theta^{t-1})$ to obtain
\begin{equation}
    \hat{y}=\mathcal{A}(\widetilde{\boldsymbol{\mathcal{R}}};\theta^{t-1}).
\end{equation}
The hard loss between the adapter prediction $\hat{y}$ and $y$ is:
\begin{equation}
    \ell_\theta=\ell(\hat{y},y).
\end{equation}
The adapter parameters are updated via gradient descent:
\begin{equation}
    \theta_k^t\gets\theta^{t-1}-\eta_\theta\nabla\ell_\theta,
\end{equation}
where $\eta_\theta$ is the adapter learning rate. During this process, personalized local knowledge is transferred to the updated adapter which is then uploaded to the server
for aggregation.

\subsection{Homogeneous Adapter Aggregation}
After receiving the local homogeneous adapters, the server aggregates them like {\tt{FedAvg}} to update the global adapter,
\begin{equation}
    \theta^t=\sum_{k=0}^{K-1}{\frac{n_k}{n}\theta_k^t}.
\end{equation}
The updated global adapter combines local knowledge across heterogeneous local models from different clients. It is then broadcast to participating clients in the next round.

\begin{figure}[t]
\centerline{\includegraphics[width=0.9\linewidth]{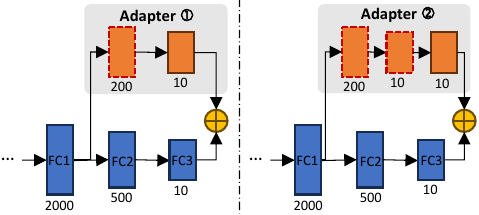}}
\vspace{-0.5em}
\caption{Two types of low-rank adapters.}
\label{fig:adapter-types}
\vspace{-1em}
\end{figure}

\subsection{Adapter Structure}
To reduce extra computational overheads by training adapters, we only match adapters for the \textit{fully connected layers} of local heterogeneous models. A low-rank adapter is an inherently ``dimension-reduced'' version of a local heterogeneous model, \emph{i.e.}, it contains far fewer parameters than the local heterogeneous model. We design two choices for constructing low-rank adapters different from typical LoRA adapters introduced in Section \ref{sec:lora-adapter}.

\textbf{Direct Dimension Reduction}. As adapter \ding{192} shown in Figure~\ref{fig:adapter-types}, we match a low-rank adapter for the last two fully connected layers of the local heterogeneous model. It consists of two linear layers: the first layer (marked with a red dashed box) is the \textit{direct dimension-reduced} version of the $FC2$ in the local heterogeneous model (dimension: $500 \rightarrow 200$), and the second layer has the same dimension as the output layer $FC3$ in the local heterogeneous model.

\textbf{Matrix Decomposition}. As adapter \ding{193} shown in Figure~\ref{fig:adapter-types}, the dimension of the parameter matrix between $FC1$ and $FC2$ in the local heterogeneous model is $2000 \times 500$. We can utilize \textit{matrix decomposition} to transform it into two small parameter matrices: $(2000 \times 200) + (200 \times 10)$ (in a $10$-class image classification task). Compared with the first adapter choice, this method reduces parameter volume while increasing network depth, which benefits from improving network learning ability. Since we only need to guarantee that the small linear LoRA adapter is a low-rank version of the fully connected layer of the large heterogeneous model, we can either manually specify the dimensions of the two decomposed matrices or leverage typical matrix decomposition approaches (\emph{e.g.} SVD) in typical LoRA adapters.

\subsection{Discussion}
In this section, we discuss the computational overheads, communication costs and privacy protection of \methodname{}.

\textbf{Computational Overhead}. On top of training a local heterogeneous model, each client also trains an extra small low-rank homogeneous adapter which contains far fewer parameters than the fully connected layers of the local heterogeneous model. Thus, the extra computational overhead by training it is acceptable.

\textbf{Communication Cost}. Each client and the FL server only exchange a small low-rank homogeneous adapter, which incurs much lower communication costs than sending a complete local model (like in {\tt{FedAvg}}).

\textbf{Privacy Protection}. Only the parameters of small low-rank homogeneous adapters are exchanged between the server and clients. Local data are always stored in clients. Hence, no private data is exposed during \methodname{} training. 


\section{Analysis}
Following \citet{FedProto,yi2023fedgh}, we first declare some additional notations. We denote $t$ as the communication round and $e \in \{0,1,\ldots,E\}$ as the iteration of local training. In each round, each client executes $E$ iterations during local training. $tE+e$ is the $e$-th iteration in the $(t+1)$-th round; $tE+0$ denotes that in the $(t+1)$-th round, before local model training, clients receive the global adapter $\mathcal{A}(\theta^t)$ aggregated in the $t$-th round; $tE+E$ is the last iteration of local training, indicating the end of local training in the $(t+1)$-th round. We also assume that the local heterogeneous model and local adapter have the same learning rate $\eta=\eta_\omega=\eta_\theta$.

\begin{assumption}\label{assump:Lipschitz}
\textbf{Lipschitz Smoothness}. The gradients of client $k$'s local heterogeneous model are $L1$--Lipschitz smooth \cite{FedProto,yi2023fedgh}, \emph{i.e.},
\vspace{-0.5em}
\begin{equation}\label{eq:7}
\footnotesize
\begin{gathered}
\|\nabla \mathcal{L}_k^{t_1}(\omega_k^{t_1} ; \boldsymbol{x}, y)-\nabla \mathcal{L}_k^{t_2}(\omega_k^{t_2} ; \boldsymbol{x}, y)\| \leqslant L_1\|\omega_k^{t_1}-\omega_k^{t_2}\|, \\
\forall t_1, t_2>0, k \in\{0,1, \ldots, N-1\},(\boldsymbol{x}, y) \in D_k.
\end{gathered}
\end{equation}
\vspace{-0.5em}
The above formulation can be expressed as:
\vspace{-0.5em}
\begin{equation}
\footnotesize
\mathcal{L}_k^{t_1}-\mathcal{L}_k^{t_2} \leqslant\langle\nabla \mathcal{L}_k^{t_2},(\omega_k^{t_1}-\omega_k^{t_2})\rangle+\frac{L_1}{2}\|\omega_k^{t_1}-\omega_k^{t_2}\|_2^2 .
\vspace{-0.5em}
\end{equation}
\end{assumption}

\begin{assumption} \label{assump:Unbiased}
\textbf{Unbiased Gradient and Bounded Variance}. The random gradient $g_{\omega,k}^t=\nabla \mathcal{L}_k^t(\omega_k^t; \mathcal{B}_k^t)$ ($\mathcal{B}$ is a batch of local data) of each client's local heterogeneous model is unbiased, and the random gradient $g_{\theta,k}^t=\nabla\mathcal{L}_k^t(\theta_k^t;\mathcal{B}_k^t)$ of each client's local adapter is also unbiased, 
\vspace{-0.5em}
\begin{equation}
\footnotesize
\begin{split}
\mathbb{E}_{\mathcal{B}_k^t \subseteq D_k}[g_{\omega,k}^t]=\nabla \mathcal{L}_k^t(\omega_k^t), \\
\mathbb{E}_{\mathcal{B}_k^t\subseteq D_k}[g_{\theta,k}^t]=\nabla\mathcal{L}_k^t(\theta_k^t),
\end{split}
\vspace{-0.5em}
\end{equation}
and the variance of $g_{\omega,k}^t$ and $g_{\theta,k}^t$ are bounded by:
\vspace{-0.5em}
\begin{equation}
\footnotesize
\begin{split}
\mathbb{E}_{\mathcal{B}_k^t \subseteq D_k}[\|\nabla \mathcal{L}_k^t(\omega_k^t ; \mathcal{B}_k^t)-\nabla \mathcal{L}_k^t(\omega_k^t)\|_2^2] \leqslant \sigma^2, \\
\mathbb{E}_{\mathcal{B}_k^t \subseteq D_k}[\|\nabla \mathcal{L}_k^t(\theta_k^t ; \mathcal{B}_k^t)-\nabla \mathcal{L}_k^t(\theta_k^t)\|_2^2] \leqslant \delta^2.
\end{split}
\end{equation}  
\vspace{-0.5em}
\end{assumption}

With these assumptions, we derive the following lemma and theorem. Their proofs can be found in Appendices~\ref{sec:proof-lemma} and \ref{sec:proof-theorem}.


\begin{lemma} \label{lemma:LocalTraining}
Based on Assumptions~\ref{assump:Lipschitz} and \ref{assump:Unbiased}, during $\{0,1,...,E\}$ local iterations of the $(t+1)$-th FL training round, the loss of an arbitrary client's local heterogeneous model is bounded by:
\vspace{-0.5em}
\begin{equation}
\footnotesize
\begin{aligned}
\mathbb{E}[\mathcal{L}_{(t+1) E}] &\leq \mathcal{L}_{t E+0}+\left({L_1 \eta^2 \mu^2}-\eta \mu\right) \sum_{e=0}^{E-1}\|\nabla \mathcal{L}_{t E+e}\|_2^2 \\
&+\frac{L_1 \eta^2(\sigma^2+\delta^2)}{2}.
\end{aligned}
\vspace{-0.5em}
\end{equation}
\end{lemma}

\begin{theorem} \label{theorem:non-convex}
\textbf{Non-convex convergence rate of pFedLoRA}. Based on the above assumptions and lemma, for an arbitrary client and any $\epsilon>0$, the following inequality holds:
\vspace{-0.5em}
\begin{equation}
\footnotesize
\begin{aligned}
\frac{1}{T} \sum_{t=0}^{T-1} \sum_{e=0}^{E-1}\|\nabla \mathcal{L}_{t E+e}\|_2^2 & \leq \frac{\frac{1}{T} \sum_{t=0}^{T-1}(\mathcal{L}_{t E+0}-\mathbb{E}[\mathcal{L}_{(t+1) E}])}{\eta \mu-{L_1 \eta^2 \mu^2}} \\
& + \frac{\frac{L_1 \eta^2(\sigma^2+\delta^2)}{2}}{\eta \mu-{L_1 \eta^2 \mu^2}} <\epsilon, \\
\text { s.t. } \eta & <\frac{2 \epsilon \mu}{L_1(\sigma^2+\delta^2+2\mu^2 \epsilon)}.
\end{aligned}
\vspace{-0.5em}
\end{equation}
\end{theorem}

Therefore, in \methodname{}, an arbitrary client's local heterogeneous model converges at a non-convex rate of $\epsilon \sim \mathcal{O}(\frac{1}{T})$.

\section{Experimental Evaluation}
In this section, we compare \methodname{} against six state-of-the-art MHPFL approaches on two real-world datasets under various experiment conditions. The experiments were conducted with Pytorch on four NVIDIA GeForce RTX 3090 GPUs with 24G memory. 

\subsection{Experiment Setup}
\textbf{Datasets}. We evaluate \methodname{} and baselines on two common image classification datasets: CIFAR-10 and CIFAR-100 \footnote{\scriptsize \url{https://www.cs.toronto.edu/\%7Ekriz/cifar.html}} \cite{cifar}. They are manually divided into non-IID datasets following the method specified in \citet{pFedHN}. For CIFAR-10, we assign only data from 2 out of the 10 classes to each client (non-IID: 2/10). For CIFAR-100, we assign only data from 10 out of the 100 classes to each client (non-IID: 10/100). Then, each client's local data are divided into the training set, the evaluation set, and the testing set following the ratio of 8:1:1. The testing set is stored locally by each client, which follows the same distribution as the local training set.

\textbf{Models}. As shown in Table~\ref{tab:model-structures} (Appendix~\ref{app:exp}), each client trains CNN models on two datasets. In model-homogeneous settings, each client has the same CNN-1 and the same adapter with two fully connected layers ($\boldsymbol{x} \rightarrow Conv1 \rightarrow Conv2 \rightarrow FC1 \rightarrow$ [\textit{direct dimension-reduced $FC2$} with $hidden\_{dim}=\{100,200,300,400,500\} \rightarrow FC3$], [$\cdot$] is the homogeneous adapter). In model-heterogeneous settings, different clients are evenly deployed with \{CNN-1,..., CNN-5\} (model id is determined by client id $k\% 5$) and the homogeneous adapter containing two fully connected layers ($\boldsymbol{x} \rightarrow Conv1 \rightarrow Conv2 \rightarrow FC1 \rightarrow FC2 \rightarrow$ [\textit{matrix-decomposed $FC2$} with $hidden\_{dim}=\{20,40,60,80\} \rightarrow FC3$], [$\cdot$] is the homogeneous adapter).

\textbf{Baselines}. We compare \methodname{} with 6 advanced baselines from three categories of MHPFL shown in Section~\ref{sec:related}: {\tt{Standalone}}, clients train local models solely; \textbf{Public-data independent knowledge distillation-based MHPFL}: {\tt{FD}} \cite{FD} and {\tt{FedProto}} \cite{FedProto}; \textbf{Mutual learning-based MHPFL}: {\tt{FML}} \cite{FML} and {\tt{FedKD}} \cite{FedKD}; \textbf{Model mixup-based MHPFL}: {\tt{LG-FedAvg}} \cite{LG-FedAvg}.


\textbf{Evaluation Metrics}. \textbf{1) Accuracy}: we measure the \textit{individual test accuracy} ($\%$) of each client's local heterogeneous model and calculate the \textit{average test accuracy} of all clients' local models. \textbf{2) Communication Cost}: We trace the number of transmitted parameters when the average model accuracy reaches the target accuracy. \textbf{3) Computation Cost}: We track the consumed computation FLOPs when the average model accuracy reaches the target accuracy.

\textbf{Training Strategy}. We tune the optimal FL settings for all methods via grid search. The epochs of local training $E \in \{1, 10\}$ and the batch size of local training $B \in \{64, 128, 256, 512\}$. The optimizer for local training is SGD with learning rate $\eta=\eta_\omega=\eta_\theta=0.01$. We also tune special hyperparameters for the baselines and report the optimal results. 
We also adjust the hyperparameters $\mu$ and $hidden\_{dim}$ to achieve the best-performance \methodname{}. To compare \methodname{} with the baselines fairly, we set the total number of communication rounds $T \in \{100, 500\}$ to ensure that all algorithms converge.

\subsection{Comparison Results}
We compare \methodname{} with baselines under \textit{model-homogeneous} (a special situation in model-heterogeneous scenarios) and \textit{model-heterogeneous} settings with varied numbers of clients $N$ and client participation fraction $C$. We set up three scenarios: $\{(N=10, C=100\%), (N=50, C=20\%), (N=100, C=10\%)\}$. For ease of comparison across the three settings, $N\times C$ is set to be the same ($10$ clients participate in each round of FL). For {\tt{FML}} and {\tt{FedKD}} under model-heterogeneous settings, we regard the smallest `CNN-5' model as the small homogeneous model.

\textbf{Average Accuracy}. The results in Tables~\ref{tab:compare-homo} and \ref{tab:compare-hetero} show that the average accuracy of all personalized heterogeneous local models in \methodname{} surpasses other baselines in both model-homogeneous and model-heterogeneous settings, and shows up to $1.26\%, 1.35\%$ accuracy improvements in model-homogeneous and model-heterogeneous settings, respectively. Figure~\ref{fig:compare-hetero-converge} (Appendix~\ref{app:exp}) shows that the average test accuracy of \methodname{} and the baselines under each $\{N, C\}$ setting specified in Table~\ref{tab:compare-hetero} varies with communication rounds. \methodname{} converges to the highest average accuracy with a lower convergence speed since an extra local adapter is required to be trained. 

\begin{table}[t]
\centering
\caption{Average accuracy for \textit{model-homogeneous} FL. $N$ is the number of clients. $C$ is the fraction of participating clients in each round. `-' denotes failure to converge.}
\vspace{-0.5em}
\label{tab:compare-homo}
\resizebox{\linewidth}{!}{%
\begin{tabular}{|l|cc|cc|cc|}
\hline
           & \multicolumn{2}{c|}{N=10, C=100\%}                                                                   & \multicolumn{2}{c|}{N=50, C=20\%}                                                                                           & \multicolumn{2}{c|}{N=100, C=10\%}                                                                  \\ \cline{2-7} 
Method     & \multicolumn{1}{c|}{CIFAR-10}                               & CIFAR-100                              & \multicolumn{1}{c|}{CIFAR-10}                               & CIFAR-100                                                     & \multicolumn{1}{c|}{CIFAR-10}                              & CIFAR-100                              \\ \hline
Standalone & \multicolumn{1}{c|}{96.35}                                  & \cellcolor[HTML]{D3D3D3}74.32                                  & \multicolumn{1}{c|}{\cellcolor[HTML]{D3D3D3}95.25}          & 62.38                                                         & \multicolumn{1}{c|}{\cellcolor[HTML]{D3D3D3}92.58}         & \cellcolor[HTML]{D3D3D3}54.93          \\
FML~\cite{FML}       & \multicolumn{1}{c|}{94.83}                                  & 70.02                                  & \multicolumn{1}{c|}{93.18}                                  & 57.56                                                         & \multicolumn{1}{c|}{87.93}                                 & 46.20                                  \\
FedKD~\cite{FedKD}      & \multicolumn{1}{c|}{94.77}                                  & 70.04                                  & \multicolumn{1}{c|}{92.93}                                  & 57.56                                                         & \multicolumn{1}{c|}{90.23}                                 & 50.99                                  \\
LG-FedAvg~\cite{LG-FedAvg}  & \multicolumn{1}{c|}{\cellcolor[HTML]{D3D3D3}96.47}          & 73.43                                  & \multicolumn{1}{c|}{94.20}                                  & 61.77                                                         & \multicolumn{1}{c|}{90.25}                                 & 46.64                                  \\
FD~\cite{FD}         & \multicolumn{1}{c|}{96.30}                                  & -                                      & \multicolumn{1}{c|}{-}                                      & -                                                             & \multicolumn{1}{c|}{-}                                     & -                                      \\
FedProto~\cite{FedProto}   & \multicolumn{1}{c|}{95.83}                                  & 72.79                                  & \multicolumn{1}{c|}{95.10}                                  & \cellcolor[HTML]{D3D3D3}62.55                                 & \multicolumn{1}{c|}{91.19}                                 & 54.01                                  \\ \hline
pFedLoRA    & \multicolumn{1}{c|}{\cellcolor[HTML]{9B9B9B}\textbf{96.69}} & \cellcolor[HTML]{9B9B9B}\textbf{75.58} & \multicolumn{1}{c|}{\cellcolor[HTML]{9B9B9B}\textbf{95.55}} & \cellcolor[HTML]{9B9B9B}{\color[HTML]{000000} \textbf{62.55}} & \multicolumn{1}{c|}{\cellcolor[HTML]{9B9B9B}\textbf{92.80}} & \cellcolor[HTML]{9B9B9B}\textbf{55.82} \\ \hline
\end{tabular}%
}
\vspace{-1em}
\end{table}

\begin{table}[t]
\centering
\caption{Average accuracy for \textit{model-heterogeneous} FL.}
\vspace{-0.5em}
\label{tab:compare-hetero}
\resizebox{\linewidth}{!}{%
\begin{tabular}{|l|cc|cc|cc|}
\hline
           & \multicolumn{2}{c|}{N=10, C=100\%}                                                                & \multicolumn{2}{c|}{N=50, C=20\%}                                                                    & \multicolumn{2}{c|}{N=100, C=10\%}                                                                   \\ \cline{2-7} 
Method     & \multicolumn{1}{c|}{CIFAR-10}                            & CIFAR-100                              & \multicolumn{1}{c|}{CIFAR-10}                               & CIFAR-100                              & \multicolumn{1}{c|}{CIFAR-10}                               & CIFAR-100                              \\ \hline
Standalone & \multicolumn{1}{c|}{\cellcolor[HTML]{D3D3D3}96.53}       & 72.53                                  & \multicolumn{1}{c|}{95.14}                                  & \cellcolor[HTML]{D3D3D3}62.71          & \multicolumn{1}{c|}{91.97}                                  & 53.04                                  \\
FML~\cite{FML}        & \multicolumn{1}{c|}{30.48}                               & 16.84                                  & \multicolumn{1}{c|}{-}                                      & 21.96                                  & \multicolumn{1}{c|}{-}                                      & 15.21                                  \\
FedKD~\cite{FedKD}      & \multicolumn{1}{c|}{80.20}                               & 53.23                                  & \multicolumn{1}{c|}{77.37}                                  & 44.27                                  & \multicolumn{1}{c|}{73.21}                                  & 37.21                                  \\
LG-FedAvg~\cite{LG-FedAvg}  & \multicolumn{1}{c|}{96.30}                               & 72.20                                  & \multicolumn{1}{c|}{94.83}                                  & 60.95                                  & \multicolumn{1}{c|}{91.27}                                  & 45.83                                  \\
FD~\cite{FD}          & \multicolumn{1}{c|}{96.21}                               & -                                      & \multicolumn{1}{c|}{-}                                      & -                                      & \multicolumn{1}{c|}{-}                                      & -                                      \\
FedProto~\cite{FedProto}   & \multicolumn{1}{c|}{96.51}                               & \cellcolor[HTML]{D3D3D3}72.59          & \multicolumn{1}{c|}{\cellcolor[HTML]{D3D3D3}95.48}          & 62.69                                  & \multicolumn{1}{c|}{\cellcolor[HTML]{D3D3D3}92.49}          & \cellcolor[HTML]{D3D3D3}53.67          \\ \hline
pFedLoRA    & \multicolumn{1}{c|}{\cellcolor[HTML]{9B9B9B}\textbf{96.66}} & \cellcolor[HTML]{9B9B9B}\textbf{73.58} & \multicolumn{1}{c|}{\cellcolor[HTML]{9B9B9B}\textbf{95.74}} & \cellcolor[HTML]{9B9B9B}\textbf{64.06} & \multicolumn{1}{c|}{\cellcolor[HTML]{9B9B9B}\textbf{92.58}} & \cellcolor[HTML]{9B9B9B}\textbf{53.95} \\ \hline
\end{tabular}%
}
\vspace{-1em}
\end{table}

\begin{figure}[t!]
\centering
\begin{minipage}[t]{0.5\linewidth}
\centering
\includegraphics[width=1.75in]{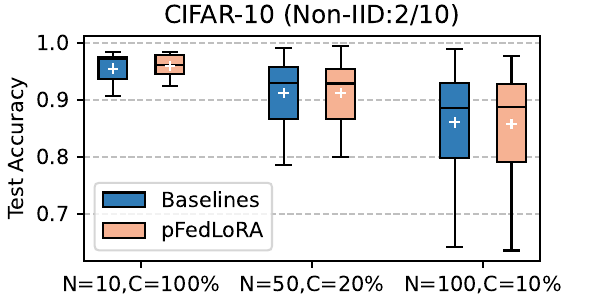}
\end{minipage}%
\begin{minipage}[t]{0.5\linewidth}
\centering
\includegraphics[width=1.75in]{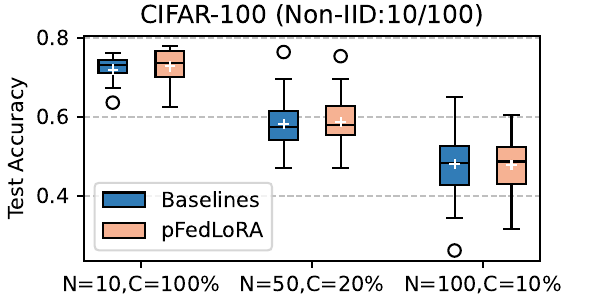}
\end{minipage}%
\vspace{-0.5em}
\caption{Accuracy distribution for individual clients.}
\label{fig:compare-hetero-individual}
\vspace{-1em}
\end{figure}

\textbf{Individual Accuracy}.
We utilize \textit{box plots} to display the distribution of individual model accuracy in model-heterogeneous settings. As shown in Figure~\ref{fig:compare-hetero-individual}, `+' denotes the average accuracy of all clients for each algorithm. A small box length bounded by the upper quartile and the lower quartile indicates a more concentrated accuracy distribution across all clients with small variance. We observe that \methodname{} obtains the higher average accuracy and the lower variance than the optimal baselines ({\tt{Standalone}} or {\tt{FedProto}} in Table~\ref{tab:compare-hetero}) at most settings.

\begin{figure}[t]
\centering
\begin{minipage}[t]{0.5\linewidth}
\centering
\includegraphics[width=1.65in]{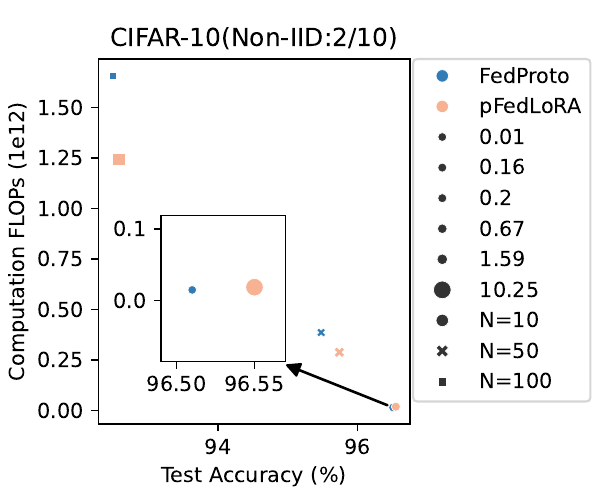}
\end{minipage}%
\begin{minipage}[t]{0.5\linewidth}
\centering
\includegraphics[width=1.65in]{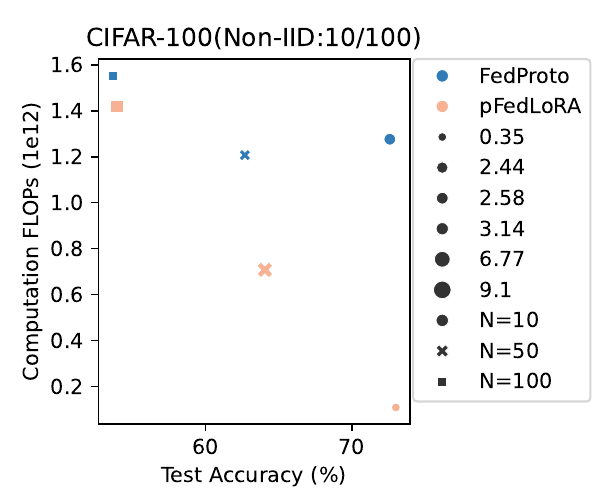}
\end{minipage}%
\caption{Trade-off among test accuracy, computational overhead and communication cost. The sizes of markers reflect the number of communicated parameters (1e6).}
\label{fig:compare-tradeoff}
\end{figure}

\textbf{Trade-off among Accuracy, Computation, Communication}. We compare \methodname{} and the state-of-the-art baseline {\tt{FedProto}} in model accuracy, computational overheads and communication costs. Figure~\ref{fig:compare-tradeoff} shows that \methodname{} always maintains the higher model accuracy and far lower computation costs
than {\tt{FedProto}} while keeping similar communication costs, indicating that \methodname{} takes the best trade-off between model accuracy, computational and communication costs. \methodname{} obtains up to $11.81 \times$ computational overhead reduction and up to $7.41\times$ communication cost saving.

\begin{figure}[t]
\centering
\begin{minipage}[t]{0.5\linewidth}
\centering
\includegraphics[width=1.65in]{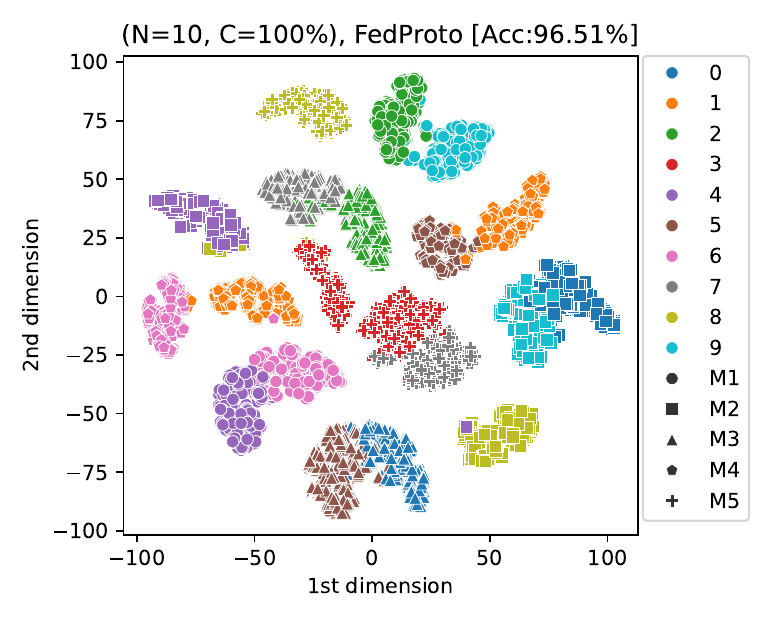}
\end{minipage}%
\begin{minipage}[t]{0.5\linewidth}
\centering
\includegraphics[width=1.65in]{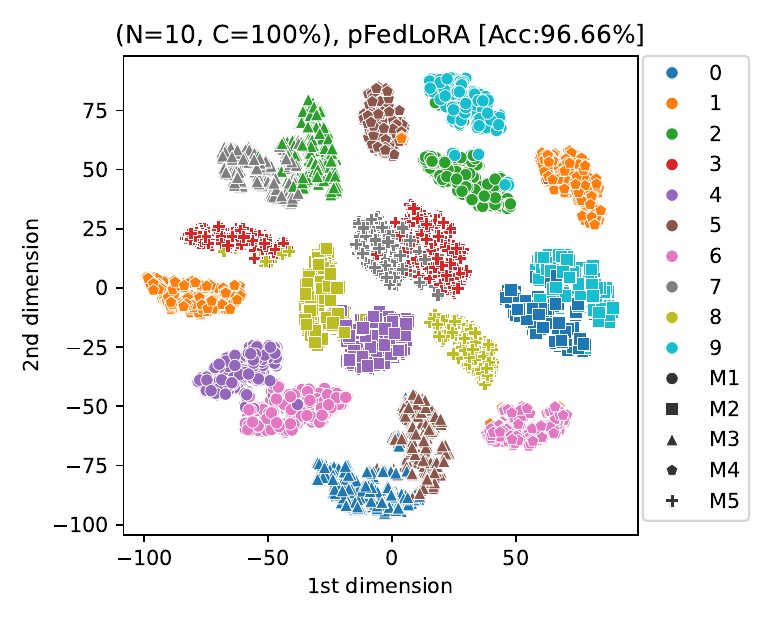}
\end{minipage}%
\caption{Representation visualization for {\tt{FedProto}} and \methodname{} on CIFAR-10 (Non-IID: 2/10).}
\label{fig:compare-TSNE}
\end{figure}

\textbf{Visualized Personalization Analysis}.
In model-heterogeneous settings, we extract every sample representation from each FL client under \methodname{} and {\tt{FedProto}}, respectively. Then, we leverage the T-SNE \cite{TSNE-JMLR} tool to reduce the dimensionality of the extracted representations from $500$ to $2$, and visualize the results. Since CIFAR-100 includes 100 classes of samples, we focus on visualizing the results on CIFAR-10 (non-IID: 2/10) in Figure~\ref{fig:compare-TSNE}. It can be observed that most clusters in \methodname{} and {\tt{FedProto}} consist of representations from a client's two seen classes of samples, which indicates that each client's local heterogeneous model has strong personalization capability. The two seen class representations within most clusters under \methodname{} and {\tt{FedProto}} satisfy ``intra-class compactness and inter-class separation'', reflecting that every client can classify its seen classes well under both algorithms. Generally, \methodname{} performs better classification boundaries than {\tt{FedProto}}.

\subsection{Case Studies}

\subsubsection{Robustness to Non-IIDness}
\begin{figure}[t]
\centering
\begin{minipage}[t]{0.5\linewidth}
\centering
\includegraphics[width=1.6in]{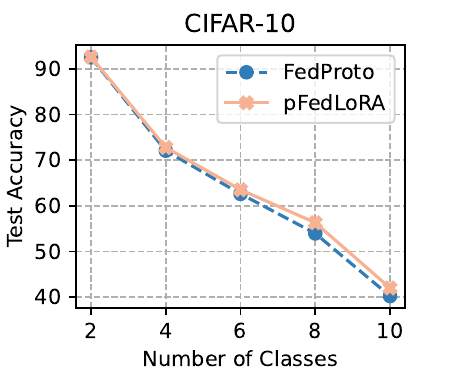}
\end{minipage}%
\begin{minipage}[t]{0.5\linewidth}
\centering
\includegraphics[width=1.6in]{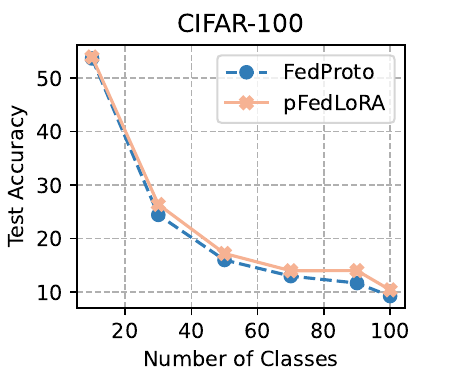}
\end{minipage}%
\caption{Robustness to Non-IIDness.}
\label{fig:case-noniid}
\end{figure}

We evaluate the robustness of \methodname{} and {\tt{FedProto}} to non-IIDness with $(N=100, C=10\%)$. We vary the number of classes seen by each client as $\{2, 4, 6, 8, 10\}$ on CIFAR-10 and $\{10, 30, 50, 70, 90, 100\}$ on CIFAR-100. Figure~\ref{fig:case-noniid} presents that \methodname{} consistently outperforms {\tt{FedProto}}, demonstrating its robustness to non-IIDness. As the non-IIDness decreases (the number of classes seen by each client rises), accuracy degrades since more IID local data enhances generalization and reduces personalization.

\subsubsection{Robustness to Client Participant Rates}
\begin{figure}[t]
\centering
\begin{minipage}[t]{0.5\linewidth}
\centering
\includegraphics[width=1.6in]{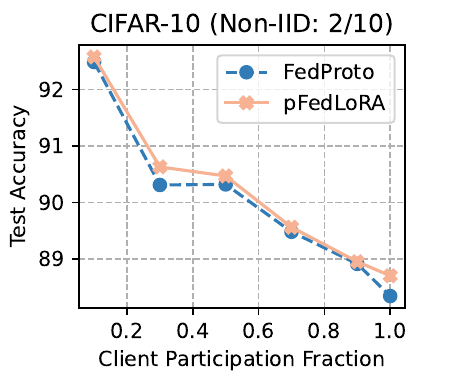}
\end{minipage}%
\begin{minipage}[t]{0.5\linewidth}
\centering
\includegraphics[width=1.6in]{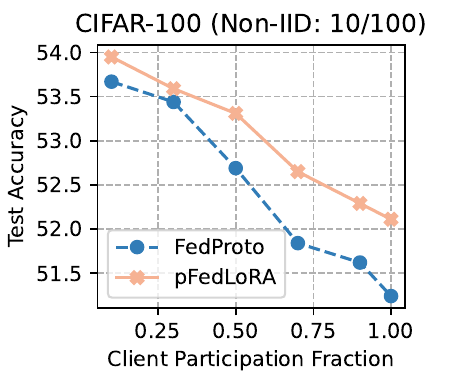}
\end{minipage}%
\caption{Robustness to client participation rates.}
\label{fig:case-frac}
\end{figure}

We also test the robustness of \methodname{} and {\tt{FedProto}} to client participant rates $C$ under $(N=100, C=10\%)$ on CIFAR-10 (non-IID: 2/10) and CIFAR-100 (non-IID: 10/100). We vary the client participant rates as $C=\{0.1, 0.3, 0.5, 0.7, 0.9, 1\}$. Figure~\ref{fig:case-frac} shows that \methodname{} consistently outperforms {\tt{FedProto}}, especially on the more complicated CIFAR-100 dataset, verifying its robustness to changes in client participant rates. Besides, as the client participant rates rise, model accuracy drops as more participating clients provide more IID local data, which also improves generalization and reduces personalization.

\section{Conclusions and Future Work}
In this paper, we propose a novel computation- and communication-efficient model-heterogeneous personalized FL framework, \methodname{}, which is inspired by LoRA tuning. It assigns a homogeneous small low-rank linear adapter for each client's local personalized heterogeneous local model. The proposed iterative learning method for training the local heterogeneous model and homogeneous adapter supports the bidirectional transfer of global knowledge and local knowledge. Aggregating the homogeneous local adapters after local iterative training on the server enables the sharing of local knowledge among FL clients. Theoretical analysis proves that \methodname{} can converge at a non-convex rate of $\mathcal{O}(\frac{1}{T})$. Extensive experiments demonstrate its superiority in model accuracy, computational overheads, and communication costs. 

In future work, we plan to explore two promising improvements for \methodname{}: a) optimizing the iterative learning process to improve model accuracy, and b) exploring lighter and more effective structures of homogeneous adapters.



\bibliographystyle{ACM-Reference-Format}
\bibliography{sample-base}


\begin{thebibliography}{49}


\ifx \showCODEN    \undefined \def \showCODEN     #1{\unskip}     \fi
\ifx \showDOI      \undefined \def \showDOI       #1{#1}\fi
\ifx \showISBNx    \undefined \def \showISBNx     #1{\unskip}     \fi
\ifx \showISBNxiii \undefined \def \showISBNxiii  #1{\unskip}     \fi
\ifx \showISSN     \undefined \def \showISSN      #1{\unskip}     \fi
\ifx \showLCCN     \undefined \def \showLCCN      #1{\unskip}     \fi
\ifx \shownote     \undefined \def \shownote      #1{#1}          \fi
\ifx \showarticletitle \undefined \def \showarticletitle #1{#1}   \fi
\ifx \showURL      \undefined \def \showURL       {\relax}        \fi
\providecommand\bibfield[2]{#2}
\providecommand\bibinfo[2]{#2}
\providecommand\natexlab[1]{#1}
\providecommand\showeprint[2][]{arXiv:#2}

\bibitem[Ahn et~al\mbox{.}(2019)]%
        {HFD1}
\bibfield{author}{\bibinfo{person}{Jin{-}Hyun Ahn} {et~al\mbox{.}}} \bibinfo{year}{2019}\natexlab{}.
\newblock \showarticletitle{Wireless Federated Distillation for Distributed Edge Learning with Heterogeneous Data}. In \bibinfo{booktitle}{\emph{Proc. {PIMRC}}}. \bibinfo{publisher}{{IEEE}}, \bibinfo{address}{Istanbul, Turkey}, \bibinfo{pages}{1--6}.
\newblock


\bibitem[Ahn et~al\mbox{.}(2020)]%
        {HFD2}
\bibfield{author}{\bibinfo{person}{Jin{-}Hyun Ahn} {et~al\mbox{.}}} \bibinfo{year}{2020}\natexlab{}.
\newblock \showarticletitle{Cooperative Learning {VIA} Federated Distillation {OVER} Fading Channels}. In \bibinfo{booktitle}{\emph{Proc. {ICASSP}}}. \bibinfo{publisher}{{IEEE}}, \bibinfo{address}{Barcelona, Spain}, \bibinfo{pages}{8856--8860}.
\newblock


\bibitem[Alam et~al\mbox{.}(2022)]%
        {FedRolex}
\bibfield{author}{\bibinfo{person}{Samiul Alam} {et~al\mbox{.}}} \bibinfo{year}{2022}\natexlab{}.
\newblock \showarticletitle{FedRolex: Model-Heterogeneous Federated Learning with Rolling Sub-Model Extraction}. In \bibinfo{booktitle}{\emph{Proc. NeurIPS}}. \bibinfo{publisher}{{}}, \bibinfo{address}{virtual}.
\newblock


\bibitem[Chang et~al\mbox{.}(2021)]%
        {Cronus}
\bibfield{author}{\bibinfo{person}{Hongyan Chang} {et~al\mbox{.}}} \bibinfo{year}{2021}\natexlab{}.
\newblock \showarticletitle{Cronus: Robust and Heterogeneous Collaborative Learning with Black-Box Knowledge Transfer}. In \bibinfo{booktitle}{\emph{Proc. NeurIPS Workshop}}. \bibinfo{publisher}{{}}, \bibinfo{address}{virtual}.
\newblock


\bibitem[Chen et~al\mbox{.}(2021)]%
        {FedMatch}
\bibfield{author}{\bibinfo{person}{Jiangui Chen} {et~al\mbox{.}}} \bibinfo{year}{2021}\natexlab{}.
\newblock \showarticletitle{FedMatch: Federated Learning Over Heterogeneous Question Answering Data}. In \bibinfo{booktitle}{\emph{Proc. {CIKM}}}. \bibinfo{publisher}{{ACM}}, \bibinfo{address}{virtual}, \bibinfo{pages}{181--190}.
\newblock


\bibitem[Cheng et~al\mbox{.}(2021)]%
        {FedGEMS}
\bibfield{author}{\bibinfo{person}{Sijie Cheng} {et~al\mbox{.}}} \bibinfo{year}{2021}\natexlab{}.
\newblock \showarticletitle{FedGEMS: Federated Learning of Larger Server Models via Selective Knowledge Fusion}.
\newblock \bibinfo{journal}{\emph{CoRR}}  \bibinfo{volume}{abs/2110.11027} (\bibinfo{year}{2021}).
\newblock


\bibitem[Cho et~al\mbox{.}(2022)]%
        {Fed-ET}
\bibfield{author}{\bibinfo{person}{Yae~Jee Cho} {et~al\mbox{.}}} \bibinfo{year}{2022}\natexlab{}.
\newblock \showarticletitle{Heterogeneous Ensemble Knowledge Transfer for Training Large Models in Federated Learning}. In \bibinfo{booktitle}{\emph{Proc. {IJCAI}}}. \bibinfo{publisher}{ijcai.org}, \bibinfo{address}{virtual}, \bibinfo{pages}{2881--2887}.
\newblock


\bibitem[Collins et~al\mbox{.}(2021)]%
        {FedRep}
\bibfield{author}{\bibinfo{person}{Liam Collins} {et~al\mbox{.}}} \bibinfo{year}{2021}\natexlab{}.
\newblock \showarticletitle{Exploiting Shared Representations for Personalized Federated Learning}. In \bibinfo{booktitle}{\emph{Proc. {ICML}}}, Vol.~\bibinfo{volume}{139}. \bibinfo{publisher}{{PMLR}}, \bibinfo{address}{virtual}, \bibinfo{pages}{2089--2099}.
\newblock


\bibitem[Diao(2021)]%
        {HeteroFL}
\bibfield{author}{\bibinfo{person}{Enmao Diao}.} \bibinfo{year}{2021}\natexlab{}.
\newblock \showarticletitle{HeteroFL: Computation and Communication Efficient Federated Learning for Heterogeneous Clients}. In \bibinfo{booktitle}{\emph{Proc. ICLR}}. \bibinfo{publisher}{OpenReview.net}, \bibinfo{address}{Virtual Event, Austria}, \bibinfo{pages}{1}.
\newblock


\bibitem[He et~al\mbox{.}(2020)]%
        {FedGKT}
\bibfield{author}{\bibinfo{person}{Chaoyang He} {et~al\mbox{.}}} \bibinfo{year}{2020}\natexlab{}.
\newblock \showarticletitle{Group Knowledge Transfer: Federated Learning of Large CNNs at the Edge}. In \bibinfo{booktitle}{\emph{Proc. NeurIPS}}. \bibinfo{publisher}{{}}, \bibinfo{address}{virtual}.
\newblock


\bibitem[Horv{\'a}th(2021)]%
        {FjORD}
\bibfield{author}{\bibinfo{person}{S. Horv{\'a}th}.} \bibinfo{year}{2021}\natexlab{}.
\newblock \showarticletitle{Fj{ORD}: Fair and Accurate Federated Learning under heterogeneous targets with Ordered Dropout}. In \bibinfo{booktitle}{\emph{Proc. {NIPS}}}. \bibinfo{publisher}{OpenReview.net}, \bibinfo{address}{Virtual}, \bibinfo{pages}{12876--12889}.
\newblock


\bibitem[Hu et~al\mbox{.}(2022)]%
        {LoRA}
\bibfield{author}{\bibinfo{person}{Edward~J. Hu} {et~al\mbox{.}}} \bibinfo{year}{2022}\natexlab{}.
\newblock \showarticletitle{LoRA: Low-Rank Adaptation of Large Language Models}. In \bibinfo{booktitle}{\emph{{ICLR}}}. \bibinfo{publisher}{OpenReview.net}, \bibinfo{address}{Virtual}, \bibinfo{pages}{1}.
\newblock


\bibitem[Huang et~al\mbox{.}(2022a)]%
        {FSFL}
\bibfield{author}{\bibinfo{person}{Wenke Huang} {et~al\mbox{.}}} \bibinfo{year}{2022}\natexlab{a}.
\newblock \showarticletitle{Few-Shot Model Agnostic Federated Learning}. In \bibinfo{booktitle}{\emph{Proc. {MM}}}. \bibinfo{publisher}{{ACM}}, \bibinfo{address}{Lisboa, Portugal}, \bibinfo{pages}{7309--7316}.
\newblock


\bibitem[Huang et~al\mbox{.}(2022b)]%
        {FCCL}
\bibfield{author}{\bibinfo{person}{Wenke Huang} {et~al\mbox{.}}} \bibinfo{year}{2022}\natexlab{b}.
\newblock \showarticletitle{Learn from Others and Be Yourself in Heterogeneous Federated Learning}. In \bibinfo{booktitle}{\emph{Proc. {CVPR}}}. \bibinfo{publisher}{{IEEE}}, \bibinfo{address}{virtual}, \bibinfo{pages}{10133--10143}.
\newblock


\bibitem[Itahara et~al\mbox{.}(2023)]%
        {DS-FL}
\bibfield{author}{\bibinfo{person}{Sohei Itahara} {et~al\mbox{.}}} \bibinfo{year}{2023}\natexlab{}.
\newblock \showarticletitle{Distillation-Based Semi-Supervised Federated Learning for Communication-Efficient Collaborative Training With Non-IID Private Data}.
\newblock \bibinfo{journal}{\emph{{IEEE} Trans. Mob. Comput.}} \bibinfo{volume}{22}, \bibinfo{number}{1} (\bibinfo{year}{2023}), \bibinfo{pages}{191--205}.
\newblock


\bibitem[Jang et~al\mbox{.}(2022)]%
        {FedClassAvg}
\bibfield{author}{\bibinfo{person}{Jaehee Jang} {et~al\mbox{.}}} \bibinfo{year}{2022}\natexlab{}.
\newblock \showarticletitle{FedClassAvg: Local Representation Learning for Personalized Federated Learning on Heterogeneous Neural Networks}. In \bibinfo{booktitle}{\emph{Proc. {ICPP}}}. \bibinfo{publisher}{{ACM}}, \bibinfo{address}{virtual}, \bibinfo{pages}{76:1--76:10}.
\newblock


\bibitem[Jeong et~al\mbox{.}(2018)]%
        {FD}
\bibfield{author}{\bibinfo{person}{Eunjeong Jeong} {et~al\mbox{.}}} \bibinfo{year}{2018}\natexlab{}.
\newblock \showarticletitle{Communication-Efficient On-Device Machine Learning: Federated Distillation and Augmentation under Non-IID Private Data}. In \bibinfo{booktitle}{\emph{Proc. NeurIPS Workshop on Machine Learning on the Phone and other Consumer Devices}}. \bibinfo{publisher}{{}}, \bibinfo{address}{virtual}.
\newblock


\bibitem[Jiang et~al\mbox{.}(2022)]%
        {PruneFL}
\bibfield{author}{\bibinfo{person}{Yuang Jiang} {et~al\mbox{.}}} \bibinfo{year}{2022}\natexlab{}.
\newblock \showarticletitle{Model Pruning Enables Efficient Federated Learning on Edge Devices}.
\newblock \bibinfo{journal}{\emph{TNNLS}} \bibinfo{volume}{1}, \bibinfo{number}{1} (\bibinfo{year}{2022}), \bibinfo{pages}{1}.
\newblock


\bibitem[Kairouz et~al\mbox{.}(2021)]%
        {1w-survey}
\bibfield{author}{\bibinfo{person}{Peter Kairouz} {et~al\mbox{.}}} \bibinfo{year}{2021}\natexlab{}.
\newblock \showarticletitle{Advances and Open Problems in Federated Learning}.
\newblock \bibinfo{journal}{\emph{Foundations and Trends in Machine Learning}} \bibinfo{volume}{14}, \bibinfo{number}{1--2} (\bibinfo{year}{2021}), \bibinfo{pages}{1--210}.
\newblock


\bibitem[Krizhevsky et~al\mbox{.}(2009)]%
        {cifar}
\bibfield{author}{\bibinfo{person}{Alex Krizhevsky} {et~al\mbox{.}}} \bibinfo{year}{2009}\natexlab{}.
\newblock \bibinfo{booktitle}{\emph{Learning multiple layers of features from tiny images}}.
\newblock \bibinfo{publisher}{Toronto, ON, Canada}, \bibinfo{address}{{}}.
\newblock


\bibitem[Li and Wang(2019)]%
        {FedMD}
\bibfield{author}{\bibinfo{person}{Daliang Li} {and} \bibinfo{person}{Junpu Wang}.} \bibinfo{year}{2019}\natexlab{}.
\newblock \showarticletitle{FedMD: Heterogenous Federated Learning via Model Distillation}. In \bibinfo{booktitle}{\emph{Proc. NeurIPS Workshop}}. \bibinfo{publisher}{{}}, \bibinfo{address}{virtual}.
\newblock


\bibitem[Li et~al\mbox{.}(2021)]%
        {FedKT}
\bibfield{author}{\bibinfo{person}{Qinbin Li} {et~al\mbox{.}}} \bibinfo{year}{2021}\natexlab{}.
\newblock \showarticletitle{Practical One-Shot Federated Learning for Cross-Silo Setting}. In \bibinfo{booktitle}{\emph{Proc. {IJCAI}}}. \bibinfo{publisher}{ijcai.org}, \bibinfo{address}{virtual}, \bibinfo{pages}{1484--1490}.
\newblock


\bibitem[Liang et~al\mbox{.}(2020)]%
        {LG-FedAvg}
\bibfield{author}{\bibinfo{person}{Paul~Pu Liang} {et~al\mbox{.}}} \bibinfo{year}{2020}\natexlab{}.
\newblock \showarticletitle{Think locally, act globally: Federated learning with local and global representations}.
\newblock \bibinfo{journal}{\emph{arXiv preprint arXiv:2001.01523}} \bibinfo{volume}{1}, \bibinfo{number}{1} (\bibinfo{year}{2020}).
\newblock


\bibitem[Lin et~al\mbox{.}(2020)]%
        {FedDF}
\bibfield{author}{\bibinfo{person}{Tao Lin} {et~al\mbox{.}}} \bibinfo{year}{2020}\natexlab{}.
\newblock \showarticletitle{Ensemble Distillation for Robust Model Fusion in Federated Learning}. In \bibinfo{booktitle}{\emph{Proc. NeurIPS}}. \bibinfo{publisher}{{}}, \bibinfo{address}{virtual}.
\newblock


\bibitem[Liu et~al\mbox{.}(2022)]%
        {CHFL}
\bibfield{author}{\bibinfo{person}{Chang Liu} {et~al\mbox{.}}} \bibinfo{year}{2022}\natexlab{}.
\newblock \showarticletitle{Completely Heterogeneous Federated Learning}.
\newblock \bibinfo{journal}{\emph{CoRR}}  \bibinfo{volume}{abs/2210.15865} (\bibinfo{year}{2022}).
\newblock


\bibitem[Lu et~al\mbox{.}(2022)]%
        {HFL}
\bibfield{author}{\bibinfo{person}{Xiaofeng Lu} {et~al\mbox{.}}} \bibinfo{year}{2022}\natexlab{}.
\newblock \showarticletitle{Heterogeneous Model Fusion Federated Learning Mechanism Based on Model Mapping}.
\newblock \bibinfo{journal}{\emph{{IEEE} Internet Things J.}} \bibinfo{volume}{9}, \bibinfo{number}{8} (\bibinfo{year}{2022}), \bibinfo{pages}{6058--6068}.
\newblock


\bibitem[Makhija et~al\mbox{.}(2022)]%
        {FedHeNN}
\bibfield{author}{\bibinfo{person}{Disha Makhija} {et~al\mbox{.}}} \bibinfo{year}{2022}\natexlab{}.
\newblock \showarticletitle{Architecture Agnostic Federated Learning for Neural Networks}. In \bibinfo{booktitle}{\emph{Proc. {ICML}}}, Vol.~\bibinfo{volume}{162}. \bibinfo{publisher}{{PMLR}}, \bibinfo{address}{virtual}, \bibinfo{pages}{14860--14870}.
\newblock


\bibitem[McMahan et~al\mbox{.}(2017)]%
        {FedAvg}
\bibfield{author}{\bibinfo{person}{Brendan McMahan} {et~al\mbox{.}}} \bibinfo{year}{2017}\natexlab{}.
\newblock \showarticletitle{Communication-Efficient Learning of Deep Networks from Decentralized Data}. In \bibinfo{booktitle}{\emph{Proc. {AISTATS}}}, Vol.~\bibinfo{volume}{54}. \bibinfo{publisher}{{PMLR}}, \bibinfo{address}{Fort Lauderdale, FL, {USA}}, \bibinfo{pages}{1273--1282}.
\newblock


\bibitem[Oh et~al\mbox{.}(2022)]%
        {FedBABU}
\bibfield{author}{\bibinfo{person}{Jaehoon Oh} {et~al\mbox{.}}} \bibinfo{year}{2022}\natexlab{}.
\newblock \showarticletitle{FedBABU: Toward Enhanced Representation for Federated Image Classification}. In \bibinfo{booktitle}{\emph{Proc. {ICLR}}}. \bibinfo{publisher}{OpenReview.net}, \bibinfo{address}{virtual}.
\newblock


\bibitem[Pillutla et~al\mbox{.}(2022)]%
        {FedAlt/FedSim}
\bibfield{author}{\bibinfo{person}{Krishna Pillutla} {et~al\mbox{.}}} \bibinfo{year}{2022}\natexlab{}.
\newblock \showarticletitle{Federated Learning with Partial Model Personalization}. In \bibinfo{booktitle}{\emph{Proc. {ICML}}}, Vol.~\bibinfo{volume}{162}. \bibinfo{publisher}{{PMLR}}, \bibinfo{address}{virtual}, \bibinfo{pages}{17716--17758}.
\newblock


\bibitem[Ruder(2016)]%
        {SGD}
\bibfield{author}{\bibinfo{person}{Sebastian Ruder}.} \bibinfo{year}{2016}\natexlab{}.
\newblock \showarticletitle{An overview of gradient descent optimization algorithms}.
\newblock \bibinfo{journal}{\emph{CoRR}}  \bibinfo{volume}{abs/1609.04747} (\bibinfo{year}{2016}), \bibinfo{pages}{1}.
\newblock


\bibitem[Sattler et~al\mbox{.}(2021)]%
        {FEDAUX}
\bibfield{author}{\bibinfo{person}{Felix Sattler} {et~al\mbox{.}}} \bibinfo{year}{2021}\natexlab{}.
\newblock \showarticletitle{FEDAUX: Leveraging Unlabeled Auxiliary Data in Federated Learning}.
\newblock \bibinfo{journal}{\emph{{IEEE} Trans. Neural Networks Learn. Syst.}} \bibinfo{volume}{1}, \bibinfo{number}{1} (\bibinfo{year}{2021}), \bibinfo{pages}{1--13}.
\newblock


\bibitem[Sattler et~al\mbox{.}(2022)]%
        {CFD}
\bibfield{author}{\bibinfo{person}{Felix Sattler} {et~al\mbox{.}}} \bibinfo{year}{2022}\natexlab{}.
\newblock \showarticletitle{{CFD:} Communication-Efficient Federated Distillation via Soft-Label Quantization and Delta Coding}.
\newblock \bibinfo{journal}{\emph{{IEEE} Trans. Netw. Sci. Eng.}} \bibinfo{volume}{9}, \bibinfo{number}{4} (\bibinfo{year}{2022}), \bibinfo{pages}{2025--2038}.
\newblock


\bibitem[Shamsian et~al\mbox{.}(2021)]%
        {pFedHN}
\bibfield{author}{\bibinfo{person}{Aviv Shamsian} {et~al\mbox{.}}} \bibinfo{year}{2021}\natexlab{}.
\newblock \showarticletitle{Personalized Federated Learning using Hypernetworks}. In \bibinfo{booktitle}{\emph{Proc. {ICML}}}, Vol.~\bibinfo{volume}{139}. \bibinfo{publisher}{{PMLR}}, \bibinfo{address}{virtual}, \bibinfo{pages}{9489--9502}.
\newblock


\bibitem[Shen et~al\mbox{.}(2020)]%
        {FML}
\bibfield{author}{\bibinfo{person}{Tao Shen} {et~al\mbox{.}}} \bibinfo{year}{2020}\natexlab{}.
\newblock \showarticletitle{Federated Mutual Learning}.
\newblock \bibinfo{journal}{\emph{CoRR}}  \bibinfo{volume}{abs/2006.16765} (\bibinfo{year}{2020}).
\newblock


\bibitem[Tan et~al\mbox{.}(2022)]%
        {FedProto}
\bibfield{author}{\bibinfo{person}{Yue Tan} {et~al\mbox{.}}} \bibinfo{year}{2022}\natexlab{}.
\newblock \showarticletitle{FedProto: Federated Prototype Learning across Heterogeneous Clients}. In \bibinfo{booktitle}{\emph{Proc. {AAAI}}}. \bibinfo{publisher}{{AAAI} Press}, \bibinfo{address}{virtual}, \bibinfo{pages}{8432--8440}.
\newblock


\bibitem[van~der Maaten and Hinton(2008)]%
        {TSNE-JMLR}
\bibfield{author}{\bibinfo{person}{Laurens van~der Maaten} {and} \bibinfo{person}{Geoffrey Hinton}.} \bibinfo{year}{2008}\natexlab{}.
\newblock \showarticletitle{Visualizing Data using t-SNE}.
\newblock \bibinfo{journal}{\emph{Journal of Machine Learning Research}} \bibinfo{volume}{9}, \bibinfo{number}{86} (\bibinfo{year}{2008}), \bibinfo{pages}{2579--2605}.
\newblock


\bibitem[W(1976)]%
        {Cauchy-Schwarz}
\bibfield{author}{\bibinfo{person}{Rudin. W}.} \bibinfo{year}{1976}\natexlab{}.
\newblock \bibinfo{booktitle}{\emph{Principles of Mathematical Analysis (3rd ed.)}}.
\newblock \bibinfo{publisher}{P McGraw-Hill., ISBN-13: 978-0070542358.}
\newblock


\bibitem[wikipedia(2023)]%
        {distributive-addtion}
\bibfield{author}{\bibinfo{person}{wikipedia}.} \bibinfo{year}{2023}\natexlab{}.
\newblock
\newblock
\newblock
\shownote{\url{https://en.wikipedia.org/wiki/Dot\_product}}.


\bibitem[Wu et~al\mbox{.}(2022)]%
        {FedKD}
\bibfield{author}{\bibinfo{person}{Chuhan Wu} {et~al\mbox{.}}} \bibinfo{year}{2022}\natexlab{}.
\newblock \showarticletitle{Communication-efficient federated learning via knowledge distillation}.
\newblock \bibinfo{journal}{\emph{Nature Communications}} \bibinfo{volume}{13}, \bibinfo{number}{1} (\bibinfo{year}{2022}), \bibinfo{pages}{2032}.
\newblock


\bibitem[Ye et~al\mbox{.}(2023)]%
        {MHPFL-survey}
\bibfield{author}{\bibinfo{person}{Mang Ye} {et~al\mbox{.}}} \bibinfo{year}{2023}\natexlab{}.
\newblock \showarticletitle{Heterogeneous Federated Learning: State-of-the-art and Research Challenges}.
\newblock \bibinfo{journal}{\emph{CoRR}}  \bibinfo{volume}{abs/2307.10616} (\bibinfo{year}{2023}), \bibinfo{pages}{1}.
\newblock


\bibitem[Yi et~al\mbox{.}(2023)]%
        {yi2023fedgh}
\bibfield{author}{\bibinfo{person}{Liping Yi}, \bibinfo{person}{Gang Wang}, \bibinfo{person}{Xiaoguang Liu}, \bibinfo{person}{Zhuan Shi}, {and} \bibinfo{person}{Han Yu}.} \bibinfo{year}{2023}\natexlab{}.
\newblock \showarticletitle{FedGH: Heterogeneous Federated Learning with Generalized Global Header}. In \bibinfo{booktitle}{\emph{Proceedings of the 31st ACM International Conference on Multimedia (ACM MM'23)}}. \bibinfo{publisher}{ACM}, \bibinfo{address}{Canada}, \bibinfo{pages}{11}.
\newblock


\bibitem[Yu et~al\mbox{.}(2021)]%
        {Fed2}
\bibfield{author}{\bibinfo{person}{Fuxun Yu} {et~al\mbox{.}}} \bibinfo{year}{2021}\natexlab{}.
\newblock \showarticletitle{Fed2: Feature-Aligned Federated Learning}. In \bibinfo{booktitle}{\emph{Proc. {KDD}}}. \bibinfo{publisher}{{ACM}}, \bibinfo{address}{virtual}, \bibinfo{pages}{2066--2074}.
\newblock


\bibitem[Yu et~al\mbox{.}(2022)]%
        {FedKEMF}
\bibfield{author}{\bibinfo{person}{Sixing Yu} {et~al\mbox{.}}} \bibinfo{year}{2022}\natexlab{}.
\newblock \showarticletitle{Resource-aware Federated Learning using Knowledge Extraction and Multi-model Fusion}.
\newblock \bibinfo{journal}{\emph{CoRR}}  \bibinfo{volume}{abs/2208.07978} (\bibinfo{year}{2022}).
\newblock


\bibitem[Zhang et~al\mbox{.}(2021)]%
        {KT-pFL}
\bibfield{author}{\bibinfo{person}{Jie Zhang} {et~al\mbox{.}}} \bibinfo{year}{2021}\natexlab{}.
\newblock \showarticletitle{Parameterized Knowledge Transfer for Personalized Federated Learning}. In \bibinfo{booktitle}{\emph{Proc. NeurIPS}}. \bibinfo{publisher}{OpenReview.net}, \bibinfo{address}{virtual}, \bibinfo{pages}{10092--10104}.
\newblock


\bibitem[Zhang et~al\mbox{.}(2022)]%
        {FedZKT}
\bibfield{author}{\bibinfo{person}{Lan Zhang} {et~al\mbox{.}}} \bibinfo{year}{2022}\natexlab{}.
\newblock \showarticletitle{FedZKT: Zero-Shot Knowledge Transfer towards Resource-Constrained Federated Learning with Heterogeneous On-Device Models}. In \bibinfo{booktitle}{\emph{Proc. {ICDCS}}}. \bibinfo{publisher}{{IEEE}}, \bibinfo{address}{virtual}, \bibinfo{pages}{928--938}.
\newblock


\bibitem[Zhang and Sabuncu(2018)]%
        {CEloss}
\bibfield{author}{\bibinfo{person}{Zhilu Zhang} {and} \bibinfo{person}{Mert~R. Sabuncu}.} \bibinfo{year}{2018}\natexlab{}.
\newblock \showarticletitle{Generalized Cross Entropy Loss for Training Deep Neural Networks with Noisy Labels}. In \bibinfo{booktitle}{\emph{Proc. {NeurIPS}}}. \bibinfo{publisher}{Curran Associates Inc.}, \bibinfo{address}{Montr{\'{e}}al, Canada}, \bibinfo{pages}{8792--8802}.
\newblock


\bibitem[Zhu et~al\mbox{.}(2021)]%
        {FedGen}
\bibfield{author}{\bibinfo{person}{Zhuangdi Zhu} {et~al\mbox{.}}} \bibinfo{year}{2021}\natexlab{}.
\newblock \showarticletitle{Data-Free Knowledge Distillation for Heterogeneous Federated Learning}. In \bibinfo{booktitle}{\emph{Proc. {ICML}}}, Vol.~\bibinfo{volume}{139}. \bibinfo{publisher}{{PMLR}}, \bibinfo{address}{virtual}, \bibinfo{pages}{12878--12889}.
\newblock


\bibitem[Zhu et~al\mbox{.}(2022)]%
        {FedResCuE}
\bibfield{author}{\bibinfo{person}{Zhuangdi Zhu} {et~al\mbox{.}}} \bibinfo{year}{2022}\natexlab{}.
\newblock \showarticletitle{Resilient and Communication Efficient Learning for Heterogeneous Federated Systems}. In \bibinfo{booktitle}{\emph{Proc. {ICML}}}, Vol.~\bibinfo{volume}{162}. \bibinfo{publisher}{{PMLR}}, \bibinfo{address}{virtual}, \bibinfo{pages}{27504--27526}.
\newblock


\end{thebibliography}

\appendix
\onecolumn

\section{Algotithm Description of \methodname{}}\label{app:alg}

\begin{algorithm}[!ht]
\caption{\methodname{}}
\label{alg:FedLoRA}
\KwInput{
$N$, total number of clients; $K$, number of selected clients in one round; $T$, total number of rounds; $\eta_\omega$, learning rate of local heterogeneous models; $\eta_\theta$, learning rate of local adapters; $\mu$, weight of local heterogeneous model loss.
}

Randomly initialize local personalized heterogeneous models $[\mathcal{F}_0(\omega_0^0),\mathcal{F}_1(\omega_1^0),\ldots,\mathcal{F}_k(\omega_k^0),\ldots,\mathcal{F}_{N-1}(\omega_{N-1}^0)]$ and the global adapter $\mathcal{A}(\theta^0)$. \\ 
\For{each round t=1,...,T-1}{
    // \textbf{Server Side}: \\
    $S^t$ $\gets$ Randomly sample $K$ clients from $N$ clients; \\
    Broadcast the global adapter $\theta^{t-1}$ to sampled $K$ clients; \\
    $\theta_k^t \gets$ \textbf{ClientUpdate}($\theta^{t-1}$); \\
    \begin{tcolorbox}[colback=ylp_color2,
                  colframe=ylp_color1,
                  width=4.3cm,
                  height=1cm,
                  arc=1mm, auto outer arc,
                  boxrule=0.5pt,
                  left=0pt,right=0pt,top=0pt,bottom=0pt,
                 ]
 
        /* Aggregate Local Adapters */ \\
        $\theta^t=\sum_{k=0}^{K-1}{\frac{n_k}{n}\theta_k^t}$.  \\
   
    \end{tcolorbox}
    
  \vspace{1em}
    // \textbf{ClientUpdate}: \\
    Receive the global adapter $\theta^{t-1}$ from the server; \\
      \For{$k\in S^t$}{
    \begin{tcolorbox}[colback=ylp_color2,
                  colframe=ylp_color1,
                  width=5.5cm,
                  height=6.8cm,
                  arc=1mm, auto outer arc,
                  boxrule=0.5pt,
                  left=0pt,right=0pt,top=0pt,bottom=0pt,
                 ]
              /* Local Iterative Training */ \\
              // Freeze Adapter, Train Model \\
               \For{$(\boldsymbol{x},y)\in D_k$}{
               $\boldsymbol{\mathcal{R}}=f_k({\boldsymbol{x};\omega}_{k,conv}^{t-1})$; \\
               $\widehat{y_1}=\mathcal{A}({\boldsymbol{\mathcal{R}};\theta}^{t-1}), \widehat{y_2}=h_k({\boldsymbol{\mathcal{R}};\omega}_{k,fc}^{t-1})$; \\
               $\ell_1=\ell(\widehat{y_1},y),\ \ell_2=\ell(\widehat{y_2},y)$; \\
               $\ell_\omega=(1-\mu)\cdot\ell_1+\mu\cdot\ell_2$; \\
               $\omega_k^t\gets\omega_k^{t-1}-\eta_\omega\nabla\ell_\omega$; 
               }
              // Freeze Adapter, Train Model \\
              \For{$(\boldsymbol{x},y)\in D_k$}{
                $\widetilde{\boldsymbol{\mathcal{R}}}=f_k({\boldsymbol{x};\omega}_{k,conv}^t)$; \\
                 $\hat{y}=\mathcal{A}({\widetilde{\boldsymbol{\mathcal{R}}};\theta}^{t-1})$; \\
                  $\ell_\theta=\ \ell(\hat{y},y)$; \\
                   $\theta_k^t\gets\theta^{t-1}-\eta_\theta\nabla\ell_\theta$; 
              }
     \end{tcolorbox}
       Upload updated local adapter $\theta_k^t$ to the server. \\
    }
}
\textbf{Return} personalized heterogeneous local models $[\mathcal{F}_0(\omega_0^{T-1}),\mathcal{F}_1(\omega_1^{T-1}),\ldots,\mathcal{F}_k(\omega_k^{T-1}),\ldots,\mathcal{F}_{N-1}(\omega_{N-1}^{T-1})]$.  
\end{algorithm}

\newpage
\section{Proof for Lemma~\ref{lemma:LocalTraining}} \label{sec:proof-lemma}

\begin{proof}
As formulated in Eq.~(\ref{eq:miu}), the local heterogeneous model of an arbitrary client $k$ is updated by
\begin{equation}\label{eq:gradient-descent}
    \omega_{t+1}=\omega_t-\eta g_{\omega,k}^t=\omega_t-\nabla(\mu\cdot\mathcal{L}_{\omega_t}+(1-\mu)\cdot\mathcal{L}_{\theta_t}).
\end{equation}

Based on Assumption~\ref{assump:Lipschitz} and Eq.~(\ref{eq:gradient-descent}), we can get
\begin{equation}
\small
\begin{aligned}
\mathcal{L}_{t E+1} & \leq \mathcal{L}_{t E+0}+\langle\nabla \mathcal{L}_{t E+0},(\omega_{t E+1}-\omega_{t E+0})\rangle+\frac{L_1}{2}\|\omega_{t E+1}-\omega_{t E+0}\|_2^2 \\
& =\mathcal{L}_{t E+0}-\eta\langle\nabla \mathcal{L}_{t E+0}, \nabla(\mu \cdot \mathcal{L}_{\omega_{t E+0}}+(1-\mu) \cdot \mathcal{L}_{\theta_{t E+0}})\rangle+\frac{L_1 \eta^2}{2}\|\nabla(\mu \cdot \mathcal{L}_{\omega_{t E+0}}+(1-\mu) \cdot \mathcal{L}_{\theta_{t E+0}})\|_2^2.
\end{aligned}
\end{equation}

Take the expectations of random variable $\xi_{tE+0}$ on both sides, we have
\begin{equation}
\small
\begin{aligned}
\mathbb{E}[\mathcal{L}_{t E+1}] & \leq \mathcal{L}_{t E+0}-\eta \mathbb{E}[\langle\nabla \mathcal{L}_{t E+0}, \nabla(\mu \cdot \mathcal{L}_{\omega_{t E+0}}+(1-\mu) \cdot \mathcal{L}_{\theta_{t E+0}})\rangle]+\frac{L_1 \eta^2}{2} \mathbb{E}[\|\nabla(\mu \cdot \mathcal{L}_{\omega_{t E+0}}+(1-\mu) \cdot \mathcal{L}_{\theta_{t E+0}})\|_2^2] \\
& \stackrel{(a)}{\leq} \mathcal{L}_{t E+0}-\eta \mathbb{E}[\langle\nabla \mathcal{L}_{t E+0}, \nabla(\mu \cdot \mathcal{L}_{\omega_{t E+0}})\rangle]+\frac{L_1 \eta^2}{2} \mathbb{E}[\|\nabla(\mu \cdot \mathcal{L}_{\omega_{t E+0}}+(1-\mu) \cdot \mathcal{L}_{\theta_{t E+0}})\|_2^2] \\
& =\mathcal{L}_{t E+0}-\eta \mu\|\nabla \mathcal{L}_{\omega_{t E+0}}\|_2^2+\frac{L_1 \eta^2}{2} \mathbb{E}[\|\nabla(\mu \cdot \mathcal{L}_{\omega_{t E+0}}+(1-\mu) \cdot \mathcal{L}_{\theta_{t E+0}})\|_2^2] \\
& \stackrel{(b)}{=} \mathcal{L}_{t E+0}-\eta \mu\|\nabla \mathcal{L}_{\omega_{t E+0}}\|_2^2+\frac{L_1 \eta^2}{2}(Var(\nabla(\mu \cdot \mathcal{L}_{\omega_{t E+0}}+(1-\mu) \cdot \mathcal{L}_{\theta_{t E+0}}))+\|\nabla(\mu \cdot \mathcal{L}_{\omega_{t E+0}}+(1-\mu) \cdot \mathcal{L}_{\theta_{t E+0}})\|_2^2) \\
& \stackrel{(c)}{\leq} \mathcal{L}_{t E+0}-\eta \mu\|\nabla \mathcal{L}_{\omega_{t E+0}}\|_2^2+\frac{L_1 \eta^2}{2}((\sigma^2+\delta^2)+\|\nabla(\mu \cdot \mathcal{L}_{\omega_{t E+0}}+(1-\mu) \cdot \mathcal{L}_{\theta_{t E+0}})\|_2^2) \\
& \stackrel{(d)}{\leq} \mathcal{L}_{t E+0}-\eta \mu\|\nabla \mathcal{L}_{\omega_{t E+0}}\|_2^2+\frac{L_1 \eta^2}{2}((\sigma^2+\delta^2)+2\|\nabla(\mu \cdot \mathcal{L}_{\omega_{t E+0}})\|_2^2) \\
& =\mathcal{L}_{t E+0}+({L_1 \eta^2 \mu^2}-\eta \mu)\|\nabla \mathcal{L}_{\omega_{t E+0}}\|_2^2+\frac{L_1 \eta^2(\sigma^2+\delta^2)}{2},
\end{aligned}
\end{equation}


where $(a)$: we simply denote that $\nabla \mathcal{L}_{t E+0}=A$, $\nabla(\mu \cdot \mathcal{L}_{\omega_{t E+0}})=B$, and $\nabla((1-\mu) \cdot \mathcal{L}_{\theta_{t E+0}})=C$. Following the additive principle of derivation, $\nabla(\mu \cdot \mathcal{L}_{\omega_{t E+0}}+(1-\mu) \cdot \mathcal{L}_{\theta_{t E+0}}) = \nabla(\mu \cdot \mathcal{L}_{\omega_{t E+0}}) + \nabla((1-\mu) \cdot \mathcal{L}_{\theta_{t E+0}}) = A + B$. So $\mathbb{E}[\langle\nabla \mathcal{L}_{t E+0}, \nabla(\mu \cdot \mathcal{L}_{\omega_{t E+0}}+(1-\mu) \cdot \mathcal{L}_{\theta_{t E+0}})\rangle] = \mathbb{E}[\langle A, B+C \rangle]$. According to distributive over vector addition \cite{distributive-addtion}, $\langle A, B+C \rangle = \langle A,B \rangle + \langle A,C \rangle$. According to the geometric interpretation of the inner product, we can obtain:
$\langle {A} \cdot {C} \rangle = |A| \cdot |C| \cdot \cos(\alpha)$, $\alpha$ is the angle between vectors $A$ and $C$, $|A|$ and $|C|$ are the norm of vectors $A$ and $C$. In the training process of two models on the same dataset on the same task, their gradient vectors $A, C$ may gradually converge to similarity, with the angle $\alpha$ between them being less than 90 degrees and ultimately approaching 0 degrees. This is because they are both guided by similar data and task objectives, gradually adjusting parameters to make the model outputs more consistent with the training data. So we can safely consider $\cos(\alpha) \geq 0$. Since norms $|A|$ and $|C|$ are positive, $\langle {A} \cdot {C} \rangle \geq 0 $. So $\langle {A} \cdot {(B+C)} \rangle - \langle {A} \cdot {B} \rangle = \langle {A} \cdot {C} \rangle \geq 0$, \emph{i.e.}, 
$\mathbb{E}[\langle\nabla \mathcal{L}_{t E+0}, \nabla(\mu \cdot \mathcal{L}_{\omega_{t E+0}}+(1-\mu) \cdot \mathcal{L}_{\theta_{t E+0}})\rangle] - \mathbb{E}[\langle\nabla \mathcal{L}_{t E+0}, \nabla(\mu \cdot \mathcal{L}_{\omega_{t E+0}})\rangle] \geq 0$. So $\mathbb{E}[\langle\nabla \mathcal{L}_{t E+0}, \nabla(\mu \cdot \mathcal{L}_{\omega_{t E+0}}+(1-\mu) \cdot \mathcal{L}_{\theta_{t E+0}})\rangle] \geq \mathbb{E}[\langle\nabla \mathcal{L}_{t E+0}, \nabla(\mu \cdot \mathcal{L}_{\omega_{t E+0}})\rangle]$, then $-\eta\mathbb{E}[\langle\nabla \mathcal{L}_{t E+0}, \nabla(\mu \cdot \mathcal{L}_{\omega_{t E+0}}+(1-\mu) \cdot \mathcal{L}_{\theta_{t E+0}})\rangle] \leq -\eta\mathbb{E}[\langle\nabla \mathcal{L}_{t E+0}, \nabla(\mu \cdot \mathcal{L}_{\omega_{t E+0}})\rangle]$. 

$(b)$ follows from $Var(x)=\mathbb{E}[x^2]-(\mathbb{E}[x]^2)$. 

$(c)$ follows from Assumption~\ref{assump:Unbiased}. 

$(d)$: we denote $B = \nabla(\mu \cdot \mathcal{L}_{\omega_{t E+0}}), C= \nabla((1-\mu) \cdot \mathcal{L}_{\theta_{t E+0}})$, we should prove that $|B+C|_{2}^{2} \leq 2|B|_{2}^{2}$.
According to the Cauchy-Schwarz Inequality, we can have 
$|B+C|^2 \leq 2|B|^2 + 2|C|^2$, which is a derivation of the Cauchy-Schwarz Inequality proved in \citet{Cauchy-Schwarz}.
Given the above inequality, since $\mu \in [0.5,1)$, as $\mu$ approaches 1, $(1-\mu)$ approaches 0, so the second term $2|C|^2=2|\nabla((1-\mu) \cdot \mathcal{L}_{\theta_{t E+0}})|^2$ can be omitted. Therefore, we can get $|B+C|^2 \leq 2|B|^2$, \emph{i.e.}, $|\nabla(\mu \cdot \mathcal{L}_{\omega_{t E+0}}+(1-\mu) \cdot \mathcal{L}_{\theta_{t E+0}})|_{2}^{2} \leq 2|\nabla(\mu \cdot \mathcal{L}_{\omega_{t E+0}})|_{2}^{2}$.

Take the expectations of the heterogeneous local model $\omega$ on both sides across $E$ local iterations, we have
\begin{equation}\label{eq:lemma}
\mathbb{E}[\mathcal{L}_{(t+1) E}] \leq \mathcal{L}_{t E+0}+({L_1 \eta^2 \mu^2}-\eta \mu) \sum_{e=0}^{E-1}\|\nabla \mathcal{L}_{t E+e}\|_2^2+\frac{L_1 \eta^2(\sigma^2+\delta^2)}{2}.
\end{equation}
\end{proof}

\section{Proof for Theorem~\ref{theorem:non-convex}}\label{sec:proof-theorem}
\begin{proof}
Eq.~(\ref{eq:lemma}) can be adjusted further as
\begin{equation}
\sum_{e=0}^{E-1}\|\nabla \mathcal{L}_{t E+e}\|_2^2 \leq \frac{\mathcal{L}_{t E+0}-\mathbb{E}[\mathcal{L}_{(t+1) E}]+\frac{L_1 \eta^2(\sigma^2+\delta^2)}{2}}{\eta \mu-{L_1 \eta^2 \mu^2}}.
\end{equation}

Take the expectations of the heterogeneous local model $\omega$ on both sides across $T$ communication rounds, we have
\begin{equation}
\frac{1}{T} \sum_{t=0}^{T-1} \sum_{e=0}^{E-1}\|\nabla \mathcal{L}_{t E+e}\|_2^2 \leq \frac{\frac{1}{T} \sum_{t=0}^{T-1}(\mathcal{L}_{t E+0}-\mathbb{E}[\mathcal{L}_{(t+1) E}])+\frac{L_1 \eta^2(\sigma^2+\delta^2)}{2}}{\eta \mu-{L_1 \eta^2 \mu^2}}.
\end{equation}

Let $\Delta=\mathcal{L}_{t=0} - \mathcal{L}^* > 0$, then $\sum_{t=0}^{T-1}(\mathcal{L}_{t E+0}-\mathbb{E}[\mathcal{L}_{(t+1) E}]) \leq \Delta$, so we have
\begin{equation}\label{eq:T}
\frac{1}{T} \sum_{t=0}^{T-1} \sum_{e=0}^{E-1}\|\nabla \mathcal{L}_{t E+e}\|_2^2 \leq \frac{\frac{\Delta}{T}+\frac{L_1 \eta^2(\sigma^2+\delta^2)}{2}}{\eta \mu-{L_1 \eta^2 \mu^2}}.
\end{equation}

If the above equation can converge to a constant $\epsilon$, \emph{i.e.},
\begin{equation}
\frac{1}{T} \sum_{t=0}^{T-1} \sum_{e=0}^{E-1}\|\nabla \mathcal{L}_{t E+e}\|_2^2 \leq \frac{\frac{\Delta}{T}+\frac{L_1 \eta^2(\sigma^2+\delta^2)}{2}}{\eta \mu-{L_1 \eta^2 \mu^2}}<\epsilon, 
\end{equation}
then
\begin{equation}
T>\frac{2 \Delta}{2\epsilon(\eta \mu-L_{1}\eta^{2}\mu^{2}) - L_{1}\eta^{2}(\sigma^2+\delta^2)}.
\end{equation}

Since $T>0, \Delta>0$, so we get
\begin{equation}
{2\epsilon(\eta \mu-L_{1}\eta^{2}\mu^{2}) - L_{1}\eta^{2}(\sigma^2+\delta^2)}>0.
\end{equation}

After solving the above inequality, we can get
\begin{equation}
\eta<\frac{2 \epsilon \mu}{L_1(\sigma^2+\delta^2+2\mu^2 \epsilon)}.
\end{equation}

Since $\epsilon,\mu,\ L_1,\ \sigma^2,\ \delta^2 > 0$ are both constants, the learning rate $\eta$ of the local heterogeneous model has solutions.

Therefore, when the learning rate of the local heterogeneous model satisfies the above condition, an arbitrary client's local heterogeneous local can converge. In addition, on the right side of Eq.~(\ref{eq:T}), except for $\frac{\Delta}{T}$, $\Delta$ and other items are both constants, so the non-convex convergence rate $\epsilon \sim \mathcal{O}(\frac{1}{T})$.
\end{proof}

\newpage
\section{More Detailed Experimental Settings and Results}\label{app:exp}

\begin{table}[h]
\centering
\caption{Structures of $5$ heterogeneous CNN models with $5 \times 5$ kernel size and $16$ or $32$ filters in convolutional layers.}
\resizebox{0.5\linewidth}{!}{%
\begin{tabular}{|l|c|c|c|c|c|}
\hline
Layer Name         & CNN-1    & CNN-2   & CNN-3   & CNN-4   & CNN-5   \\ \hline
Conv1              & 5$\times$5, 16   & 5$\times$5, 16  & 5$\times$5, 16  & 5$\times$5, 16  & 5$\times$5, 16  \\
Maxpool1              & 2$\times$2   & 2$\times$2  & 2$\times$2  & 2$\times$2  & 2$\times$2  \\
Conv2              & 5$\times$5, 32   & 5$\times$5, 16  & 5$\times$5, 32  & 5$\times$5, 32  & 5$\times$5, 32  \\
Maxpool2              & 2$\times$2   & 2$\times$2  & 2$\times$2  & 2$\times$2  & 2$\times$2  \\
FC1                & 2000     & 2000    & 1000    & 800     & 500     \\
FC2                & 500      & 500     & 500     & 500     & 500     \\
FC3                & 10/100   & 10/100  & 10/100  & 10/100  & 10/100  \\ \hline
model size & 10.00 MB & 6.92 MB & 5.04 MB & 3.81 MB & 2.55 MB \\ \hline
\end{tabular}%
}
\label{tab:model-structures}
\end{table}

\begin{figure*}[h]
\centering
\begin{minipage}[t]{0.3\linewidth}
\centering
\includegraphics[width=2in]{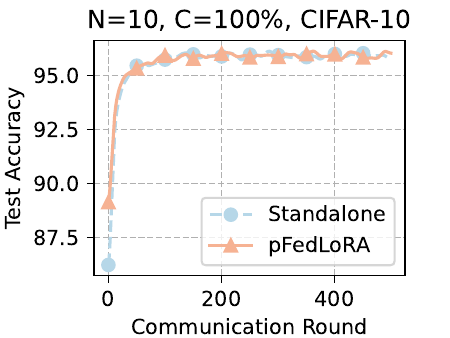}
\end{minipage}%
\begin{minipage}[t]{0.3\linewidth}
\centering
\includegraphics[width=2in]{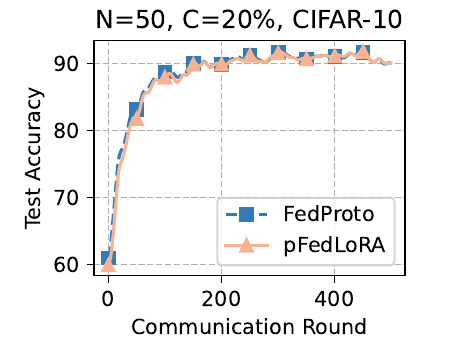}
\end{minipage}%
\begin{minipage}[t]{0.3\linewidth}
\centering
\includegraphics[width=2in]{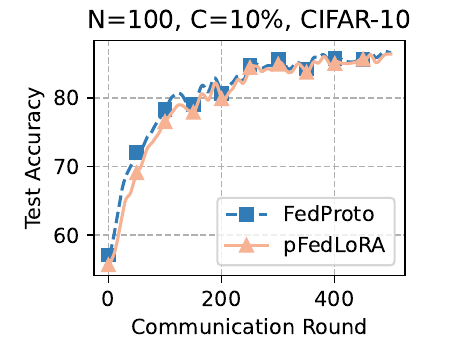}
\end{minipage}%

\begin{minipage}[t]{0.3\linewidth}
\centering
\includegraphics[width=2in]{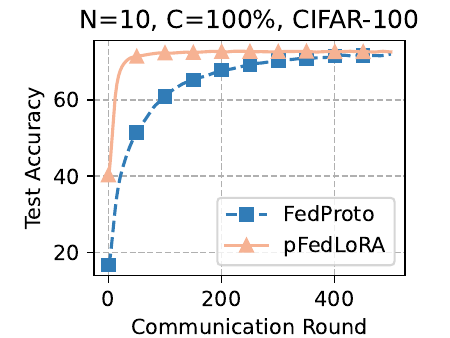}
\end{minipage}%
\begin{minipage}[t]{0.3\linewidth}
\centering
\includegraphics[width=2in]{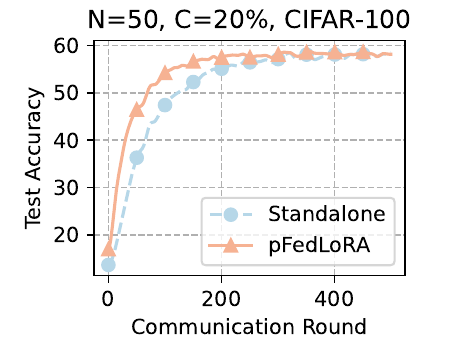}
\end{minipage}%
\begin{minipage}[t]{0.3\linewidth}
\centering
\includegraphics[width=2in]{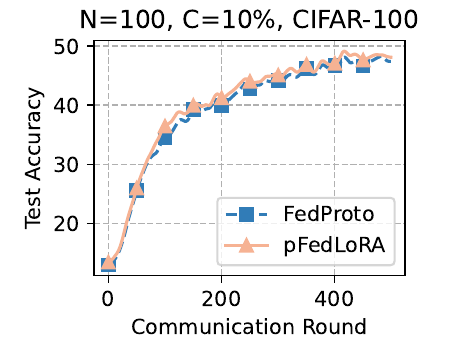}
\end{minipage}%
\caption{Average accuracy vs. communication rounds.}
\label{fig:compare-hetero-converge}
\end{figure*}

\end{document}